\documentclass{egblank}
\usepackage{eg2026}
 
\usepackage[T1]{fontenc}
\usepackage{dfadobe}  

\usepackage{cite}  %
\BibtexOrBiblatex
\electronicVersion
\PrintedOrElectronic
\ifpdf \usepackage[pdftex]{graphicx} \pdfcompresslevel=9
\else \usepackage[dvips]{graphicx} \fi
\usepackage{subcaption}
\usepackage[labelfont=bf, textfont=it]{caption}
\usepackage{egweblnk} 

\usepackage{enumitem}
\setlist{nosep} %

\usepackage[compact]{titlesec}
\titlespacing{\section}{0pt}{2ex}{1ex}
\titlespacing{\subsection}{0pt}{1ex}{0ex}
\titlespacing{\subsubsection}{0pt}{0.5ex}{0ex}

\usepackage{microtype} %

\usepackage{lipsum}

\include{svg}
\title[Beyond Segmentation: Structurally Informed Facade Parsing from Imperfect Images]%
      {Beyond Segmentation: Structurally Informed Facade Parsing from Imperfect Images}

\usepackage[normalem]{ulem} %

\newif\ifrevisions
\revisionsfalse

\DeclareRobustCommand{\rev}[1]{%
  \ifrevisions\textcolor{green}{#1}\else#1\fi
}

\DeclareRobustCommand{\del}[1]{%
  \ifrevisions\textcolor{red}{\sout{#1}}\fi
}

\author[]
{\parbox{\textwidth}{\centering M. Janicki\orcid{0009-0003-0103-5600}
         and A. Plocharski\orcid{0000-0002-7487-8153} 
         and P. Musialski\orcid{0000-0001-6429-8190}
}}
\author[M. Janicki \& A. Plocharski \& P. Musialski]
{\parbox{\textwidth}{\centering M. Janicki$^{1}$\thanks{Equal contribution}\orcid{0009-0003-0103-5600}, A. Plocharski$^{1,2}$\footnotemark[1]\orcid{0000-0002-7487-8153} and P. Musialski$^{3}$\orcid{0000-0001-6429-8190}
        }
        \\
{\parbox{\textwidth}{\centering $^1$Warsaw University of Technology, $^2$Akces NCBR, $^3$New Jersey Institute of Technology
       }
}
}

\begin{document}

\teaser{
 \includegraphics[width=1.0\linewidth]{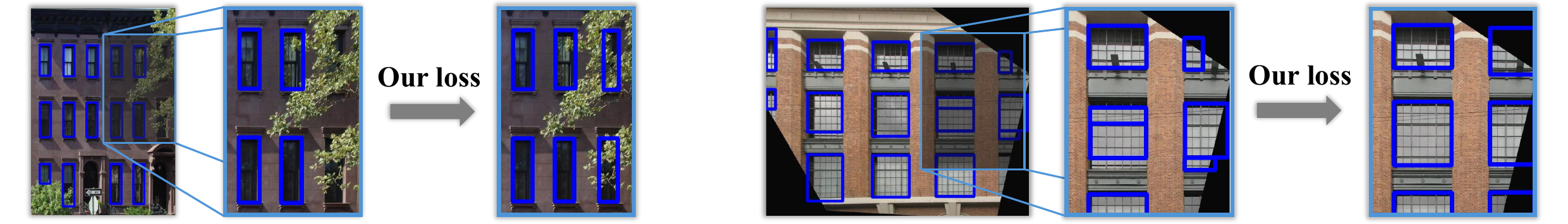}
 \centering
  \caption{Our loss term corrects errors in facade parsing resulting from image imperfections. It corrects facade elements' sizes and positions and is able to reconstruct elements that are partially obstructed.}
\label{fig:teaser}
}

\maketitle
\begin{abstract}
Standard object detectors typically treat architectural elements independently, often resulting in facade parsings that lack the structural coherence required for downstream procedural reconstruction. We address this limitation by augmenting the YOLOv8 training objective with a custom lightweight alignment loss. This regularization encourages grid-consistent arrangements of bounding boxes during training, effectively injecting geometric priors without altering the standard inference pipeline. Experiments on the CMP dataset demonstrate that our method successfully improves structural regularity, correcting alignment errors caused by perspective and occlusion while maintaining a controllable trade-off with standard detection accuracy.

\ccsdesc[500]{Computing methodologies~Image segmentation}
\ccsdesc[500]{Computing methodologies~Neural networks}
\ccsdesc[300]{Applied computing~Architecture (buildings)}

\printccsdesc   
\end{abstract}

\section{Introduction}
\vspace{-2mm}

Reconstructing architectural facade structure from a single image is a core problem in urban modeling, with applications in digital twins, scanning-to-CAD workflows, and content creation. Recent neuro-symbolic methods, such as FaçAID~\cite{plocharski2024facaid}, can convert facade segmentations into editable procedural programs. In practice, however, these pipelines are bottlenecked by the structure of the visual input: they work best when detected elements form a coherent, regular arrangement.

Modern detectors (e.g., YOLOv8, Mask R-CNN) are optimized for local appearance cues and treat elements largely independently. On facades, this commonly produces outputs that are visually plausible yet structurally inconsistent: small boundary jitter across repeated elements, gaps under occlusion, and systematic drifts under residual perspective or relief effects. When such detections are used as input to procedural reconstruction, the downstream model must absorb the inconsistency, typically increasing ambiguity and complexity of the inferred structure.

We address this issue upstream, at the detector level. We augment a standard YOLOv8 training objective with a lightweight pairwise alignment regularizer that encourages consistent row/column structure among same-class predictions. The intent is not to impose an explicit grammar, but to bias training toward grid-consistent detections while keeping inference unchanged.

\noindent
\del{\textbf{Contributions.} }We contribute:
\begin{itemize}
    \item a detector-side training objective that injects facade-specific geometric regularity into bounding-box predictions via a pairwise alignment term;
    \item an integration into a standard YOLOv8 training pipeline that preserves the original inference procedure;
    \item \del{an empirical evaluation on CMP demonstrating improved structural regularity of detections relative to the unregularized baseline, and characterizing the accuracy--regularity trade-off}\rev{a weight parameter that controls the trade-off between detection accuracy and structural regularity}.
\end{itemize}

\vspace{-1mm}
\section{Related Work}
\vspace{-2mm}

\noindent
\textbf{Facade Detection and Segmentation.} Early approaches to facade parsing relied on strong priors or split grammars to guide segmentation~\cite{teboul2010segmentation, simon2011procedural}. While structurally robust, these methods were computationally expensive and struggled to generalize to diverse architectural styles. The advent of deep learning shifted the focus to pure semantic segmentation~\cite{liu2020deepfacade} and object detection~\cite{yolov8}. These methods offer real-time inference and high recall but treat architectural elements as independent objects, ignoring the global regularities (alignment, repetition) that define facade structures. Prior work addresses occlusion and irregularity through explicit repair based on symmetry~\cite{musialski2009symmetry} or statistical structure completion~\cite{fan2014structure}; in contrast, our method embeds these geometric priors directly into the detection pipeline.

\noindent
\textbf{Inverse Procedural Modeling (IPM).} IPM methods aim to recover procedural rules from images or 3D data~\cite{musialski2012inverse, fan2014inverse}. Recent works like FaçAID~\cite{plocharski2024facaid} use transformer-based architectures to infer grammar parameters from segmented inputs. However, these systems are typically sensitive to input noise; they function best when the input segmentation already reflects the logical structure of the building. Our work complements this direction by ensuring the input detections are geometrically consistent, thereby stabilizing the inverse modeling process.

\vspace{-1mm}
\section{Method}
\vspace{-2mm}

\noindent
\textbf{YOLOv8. } 
The YOLOv8 model serves as the starting point for our method. YOLOv8 is a state-of-the-art, single-stage object detector. Unlike previous anchor-based iterations, YOLOv8 utilizes an \emph{anchor-free} detection head that predicts object centers and dimensions directly from grid points in the feature map.

During training, YOLOv8 employs a dynamic task-aligned assigner to select positive samples that best match ground truth objects. These selected positive bounding boxes \del{serve as }\rev{are} the input to our \del{alignment }loss. The baseline YOLO objective consists of three types of loss terms:
\begin{itemize}
    \item \emph{Complete Intersection over Union (CIoU)} bounding box loss,
    \item \emph{Binary Cross Entropy (BCE)} for classification,
    \item \emph{Distribution Focal Loss (DFL)} for precise box regression.
\end{itemize}

\noindent
During inference, YOLO uses Non-Maximum Suppression (NMS) to remove duplicate detections. In our experiments, we set the confidence threshold to 0.25.

\noindent
\textbf{Alignment loss term. } 
Standard object detectors process detections independently, often ignoring the global regularity. To address this, we introduce a custom loss term that mitigates the influence of image imperfections (e.g., perspective distortion or obstructions) by enforcing better alignment between predicted boxes. This term is added to the baseline YOLOv8 objective, balanced by a weight $W$.

Our loss term operates on the same set of positive bounding boxes selected by the YOLO assigner. The boxes are defined in pixel coordinates as tuples $(x_1, y_1, x_2, y_2)$. The optimization goal is to align bounding boxes in both axes. Here, we describe the definition for $x$-axis alignment; the $y$-axis definition is analogous.

\noindent
We identify a pair of bounding boxes as a candidate for alignment if they meet the following conditions:
\begin{itemize}
\item both bounding boxes are of the same class,
\item bounding boxes \del{don't}\rev{do not} overlap (IoU $\approx$ 0),
\item the abs difference between $x_1$ coordinates is below threshold $T$,
\item the abs difference between $x_2$ coordinates is below threshold $T$.
\end{itemize}
\noindent
Threshold $T$ \rev{(defined in pixels) }is a hyperparameter. If all conditions are met for a pair with indices $i$ and $j$, the loss is:
$$
L_{ij}^x = |x_1^i - x_1^j| + |x_2^i - x_2^j|
$$
where $x$ denotes the axis. If conditions are not met, the loss is $0$.
\noindent
Finally, the total alignment loss is calculated as the normalized sum over all valid pairs:
$$
L = \frac{\sum_{i<j} L_{ij}^x}{\max(n^x, 1)} + \frac{\sum_{i<j} L_{ij}^y}{\max(n^y, 1)}
$$
where $n^x$ and $n^y$ are the number of pairs meeting conditions for the $x$-axis and $y$-axis, respectively. This term encourages spatially proximate boxes to align to a common grid.

\begin{figure}[t]
  \centering
  \includegraphics[width=.7\linewidth]{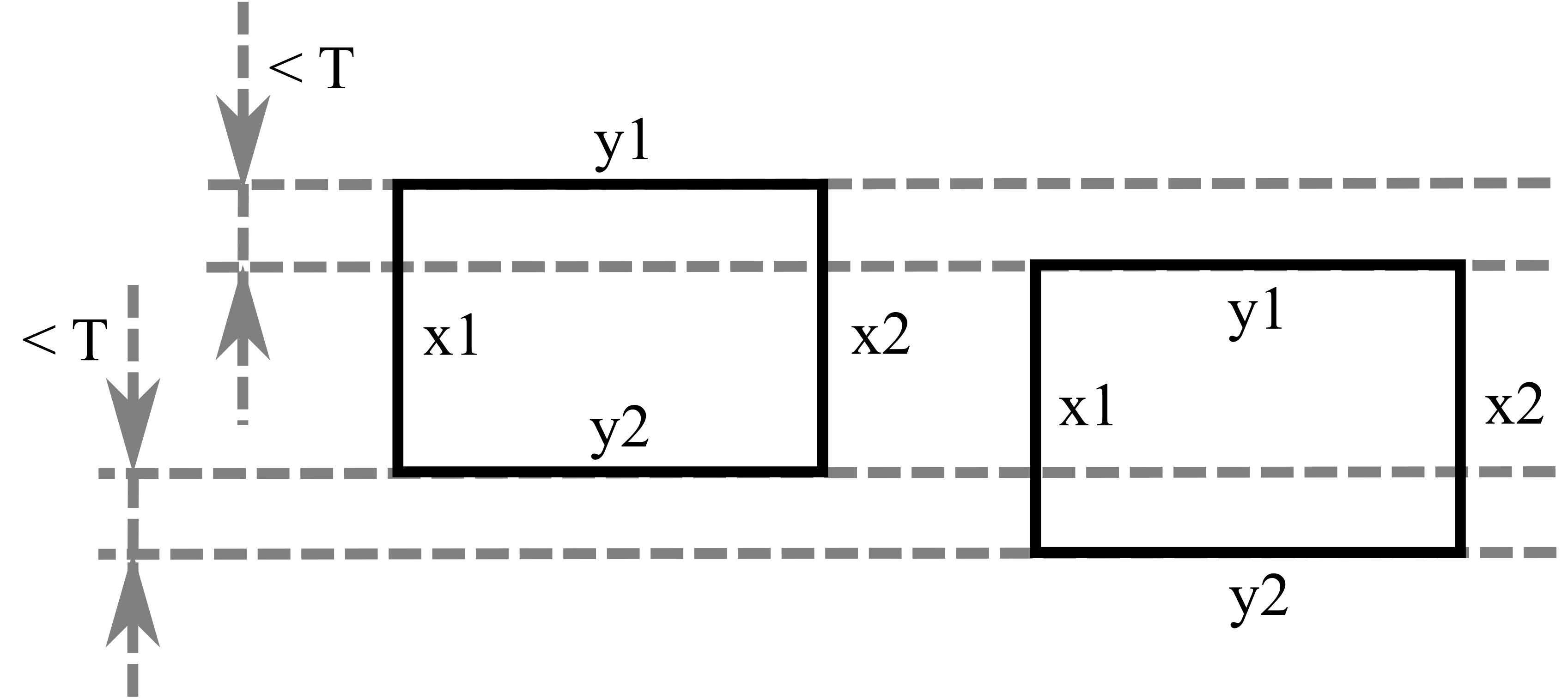}
  \caption{\label{fig:bboxes}
           Example of a pair of bounding boxes meeting the required conditions (for y-axis): coordinate differences are below threshold and the boxes \del{don't}\rev{do not} overlap.}
\end{figure}

\begin{figure*}[t]
  \centering
  \begin{subfigure}[b]{0.32\linewidth}
    \includegraphics[width=\linewidth]{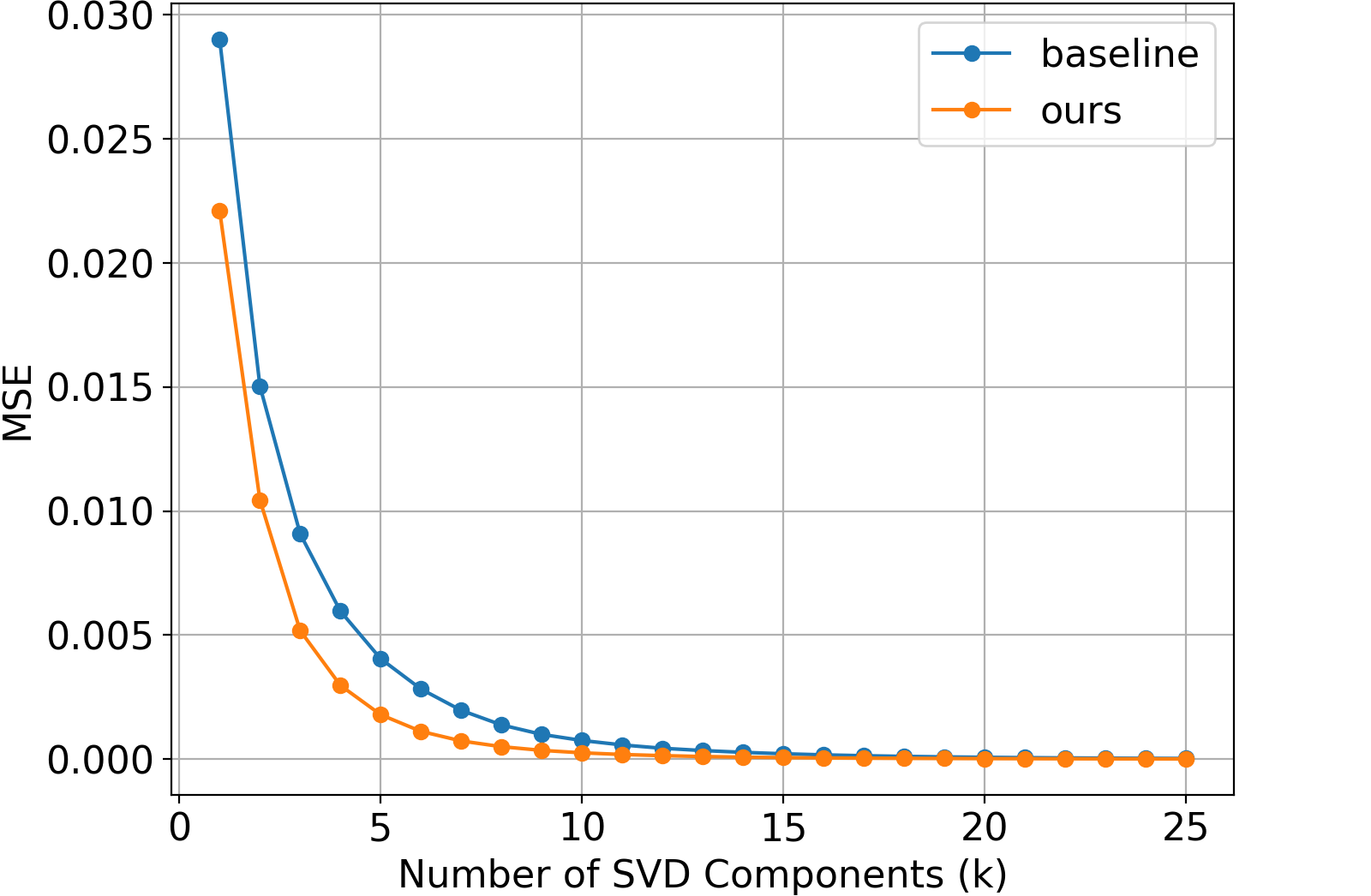}
    \caption{MSE of rank-$k$ approximations}
    \label{fig:MSE}
  \end{subfigure}
  \hfill
  \begin{subfigure}[b]{0.32\linewidth}
    \includegraphics[width=\linewidth]{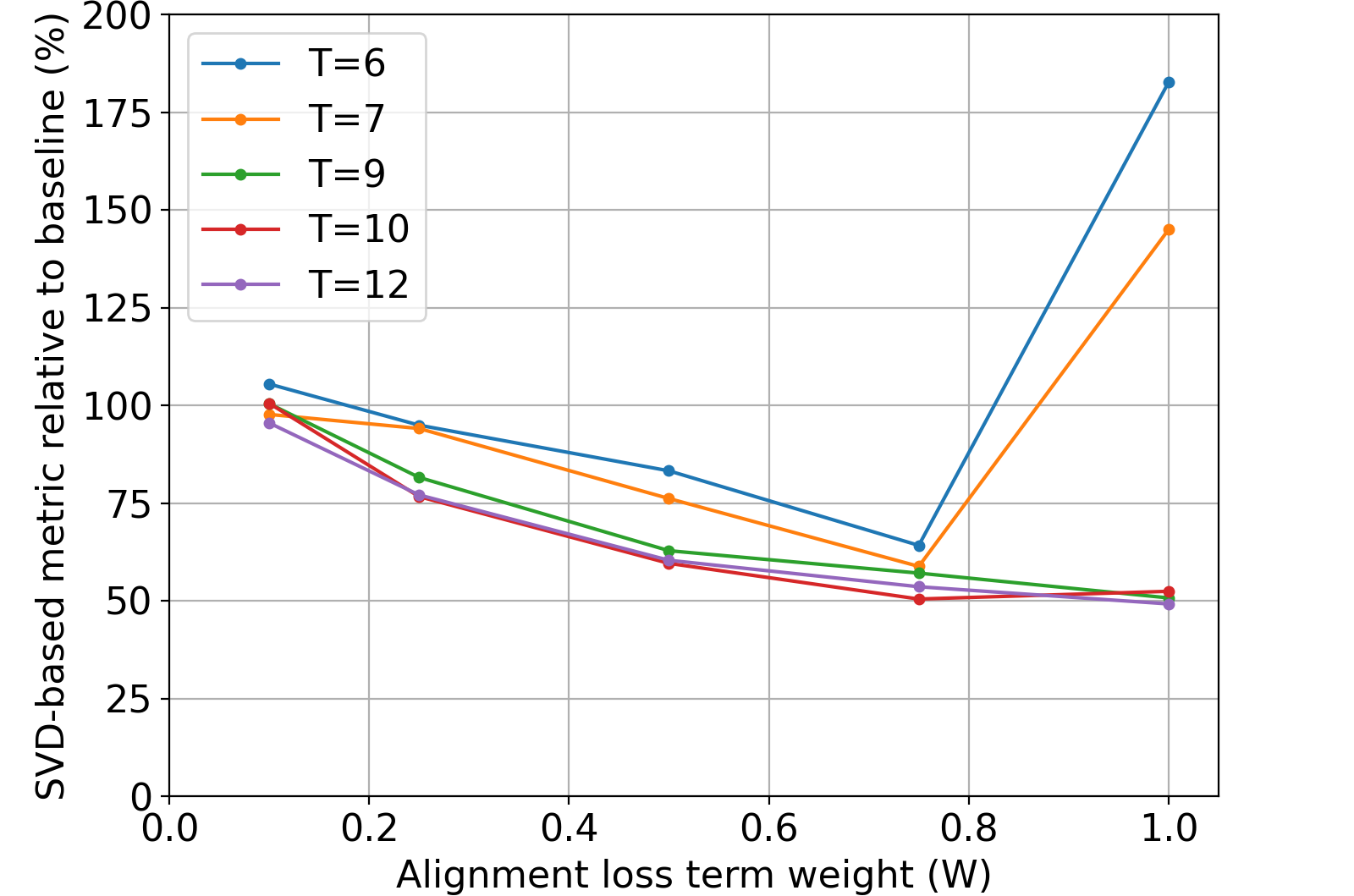}
    \caption{SVD metric vs. Weight ($W$) and Threshold ($T$)}
    \label{fig:SVD_W}
  \end{subfigure}
  \hfill
  \begin{subfigure}[b]{0.32\linewidth}
    \includegraphics[width=\linewidth]{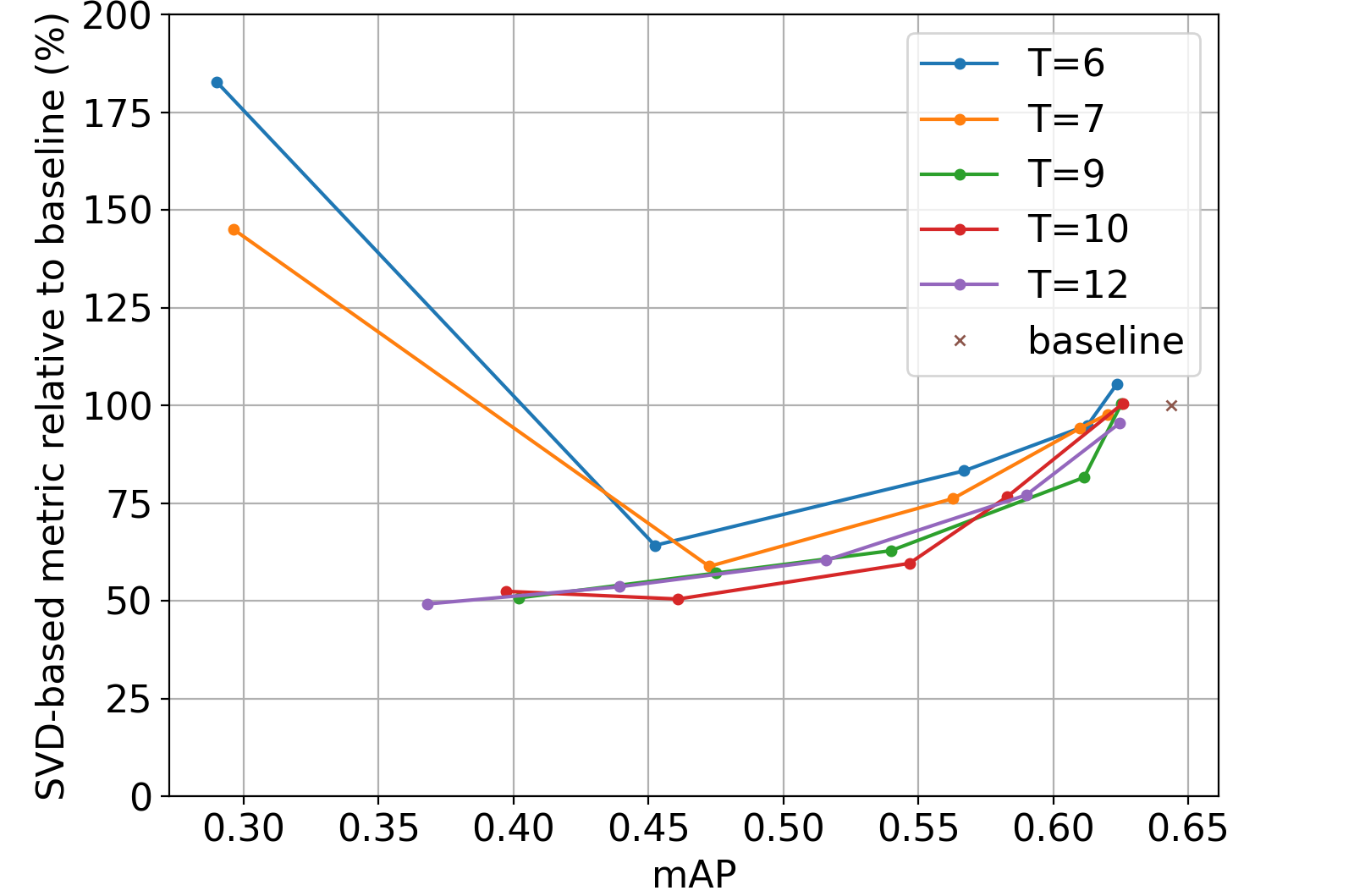}
    \caption{Trade-off: SVD metric vs. mAP}
    \label{fig:SVD_mAP}
  \end{subfigure}
  \caption{Analysis of structural regularity. (a) Mean Squared Error (MSE) of the rank-$k$ approximation for windows; lower curves indicate better low-rank approximation (higher regularity). (b) Relative SVD-based regularity score for varying alignment weights $W$ and thresholds $T$ (baseline = 100\%; lower is better). (c) The trade-off between structural regularity (SVD metric, lower is better) and detection accuracy (mAP, higher is better), showing that moderate alignment weights improve regularity with minimal impact on mAP.}
  \label{fig:analysis_plots}
\end{figure*}

\vspace{-1mm}
\section{Experiments}
\vspace{-2mm}

\noindent
\textbf{Dataset.}
We train and evaluate our model on the CMP facade dataset~\cite{Tylecek13}, selected for its high-quality pixel-level annotations. The dataset contains 606 annotated images comprising 689 individual facades. We crop these images to the bounds of individual facades, resulting in 689 samples. The data is randomly split into training (80\%), validation (10\%), and test (10\%) sets.

\noindent
\textbf{Training.}
\rev{To establish a baseline, we fine-tuned pretrained YOLOv8 Small and YOLOv11 Small models for 200 epochs. The best checkpoints were selected based on the highest mAP@0.5 (Mean Average Precision) score on the validation set. The relative difference in that score between YOLOv8 and YOLOv11 was about $0.1\%$. We also observed no significant difference in speed of training or inference. Thus, since our codebase was based around YOLOv8's code, we decided to keep it as the only YOLO baseline for improved results clarity.} \del{We establish a baseline by fine-tuning a pretrained YOLOv8 Small model on the training set for 200 epochs. The best baseline checkpoint is selected based on the highest mAP@0.5 score on the validation set. }Our method follows the exact same initialization and training schedule \rev{as the baseline} but incorporates the alignment loss ($L_{align}$) with weight $W$. We perform an ablation study across multiple values of threshold $T$ \del{(pixels) }and weight $W$ to analyze their impact.\rev{ All tested models can be found on \href{https://beyond-segmentation.github.io}{https://beyond-segmentation.github.io}.}

\noindent
\textbf{Evaluation Metrics.}
We employ two distinct metrics to capture the trade-off between visual fidelity and structural regularity:

\textbf{(1) mAP@0.5:} We use the standard Mean Average Precision (mAP) at an IoU threshold of 0.5. While the CMP dataset is rectified, residual perspective distortions and depth-induced parallax (e.g., balconies vs. windows) often cause ground truth boxes to deviate slightly from a perfect grid. Since our method aims to correct these visual offsets to architectural truth, a slight drop in standard mAP is expected and acceptable if it favors structural consistency.

\textbf{(2) SVD-based Regularity:} To quantify the structural coherence of the predictions, we adopt the Singular Value Decomposition (SVD) metric proposed by Plocharski et al.~\cite{plocharski2025prodg}, which builds upon the rank-one approximation concepts introduced by Yang et al.~\cite{yang2012parsing}. The core insight is that a highly regular, grid-aligned structure has a lower algebraic rank than a disordered one. 

We compute this metric on the binary masks of detected windows. The mask is decomposed via SVD and approximated using the sum of its first $k$ rank-1 components. The regularity score is defined as the sum of Mean Squared Errors (MSE) between the original mask and its rank-$k$ approximations for $k \in [1, 25]$. A lower score indicates that the facade structure is captured by fewer components, implying higher geometric regularity (see Figure~\ref{fig:MSE}).

\subsection{Results}
\vspace{-1mm}

\begin{figure*}[!t]
\centering
\def\fixedheight{1.38cm} 

\setlength{\tabcolsep}{1pt} %

\begin{minipage}[t]{0.32\textwidth}
  \centering
  \begin{tabular}{ccc}
    \small\textbf{GT} & \small\textbf{Baseline} & \small\textbf{Ours} \\
    \includegraphics[height=\fixedheight, width=0.32\linewidth, keepaspectratio]{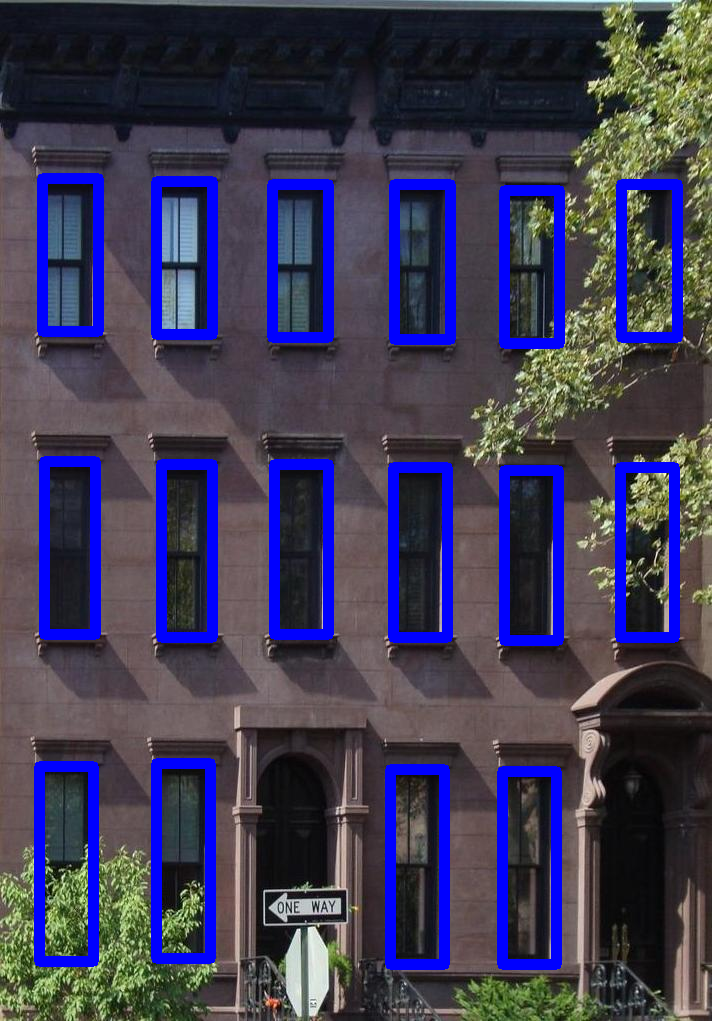} &
    \includegraphics[height=\fixedheight, width=0.32\linewidth, keepaspectratio]{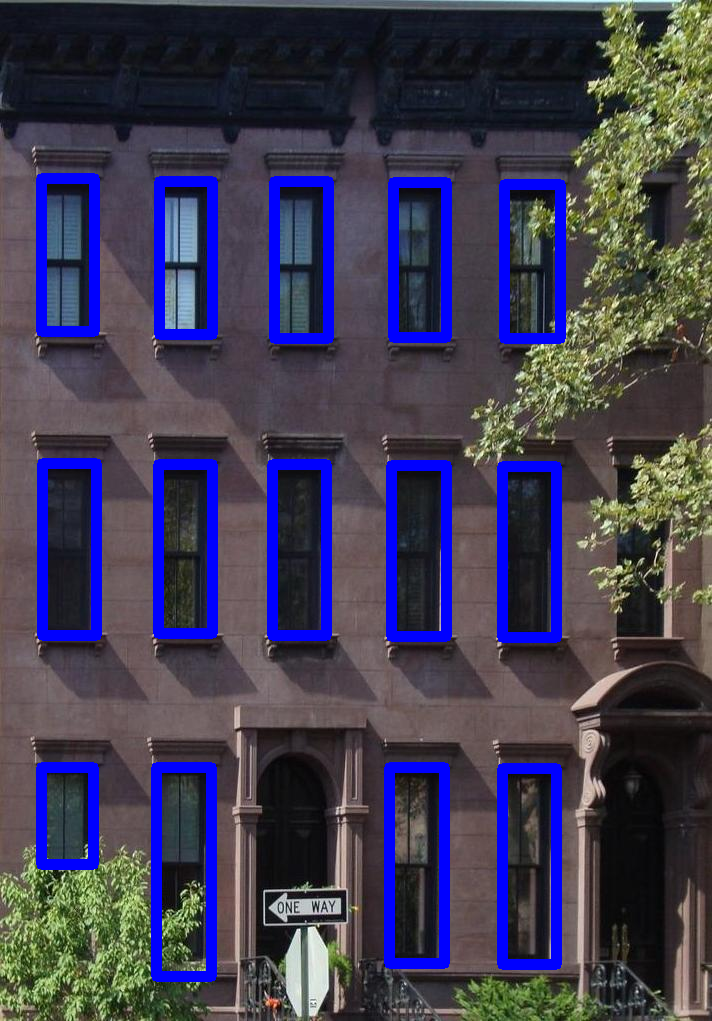} &
    \includegraphics[height=\fixedheight, width=0.32\linewidth, keepaspectratio]{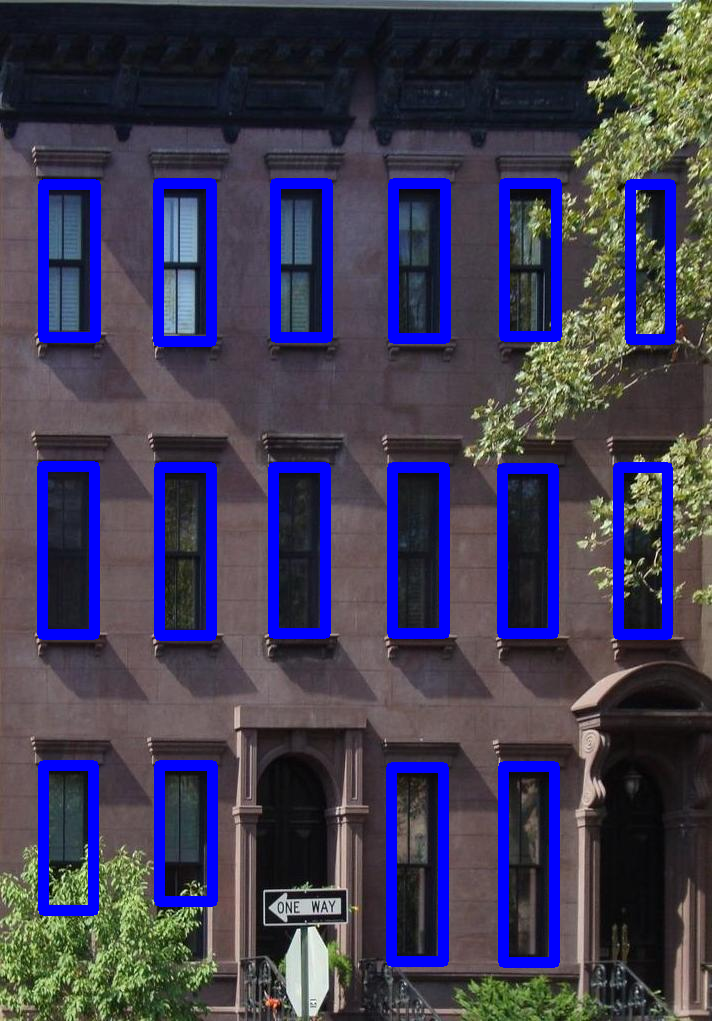} \\
    
    \includegraphics[height=\fixedheight, width=0.32\linewidth, keepaspectratio]{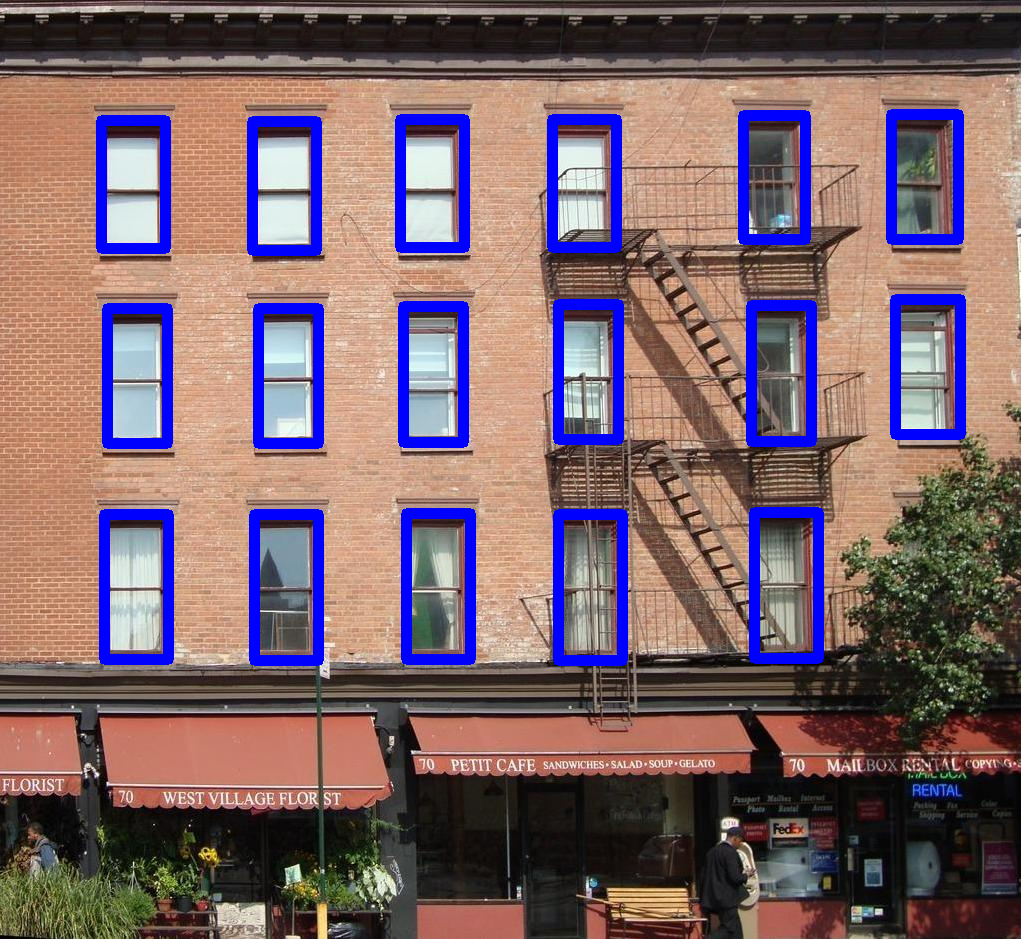} &
    \includegraphics[height=\fixedheight, width=0.32\linewidth, keepaspectratio]{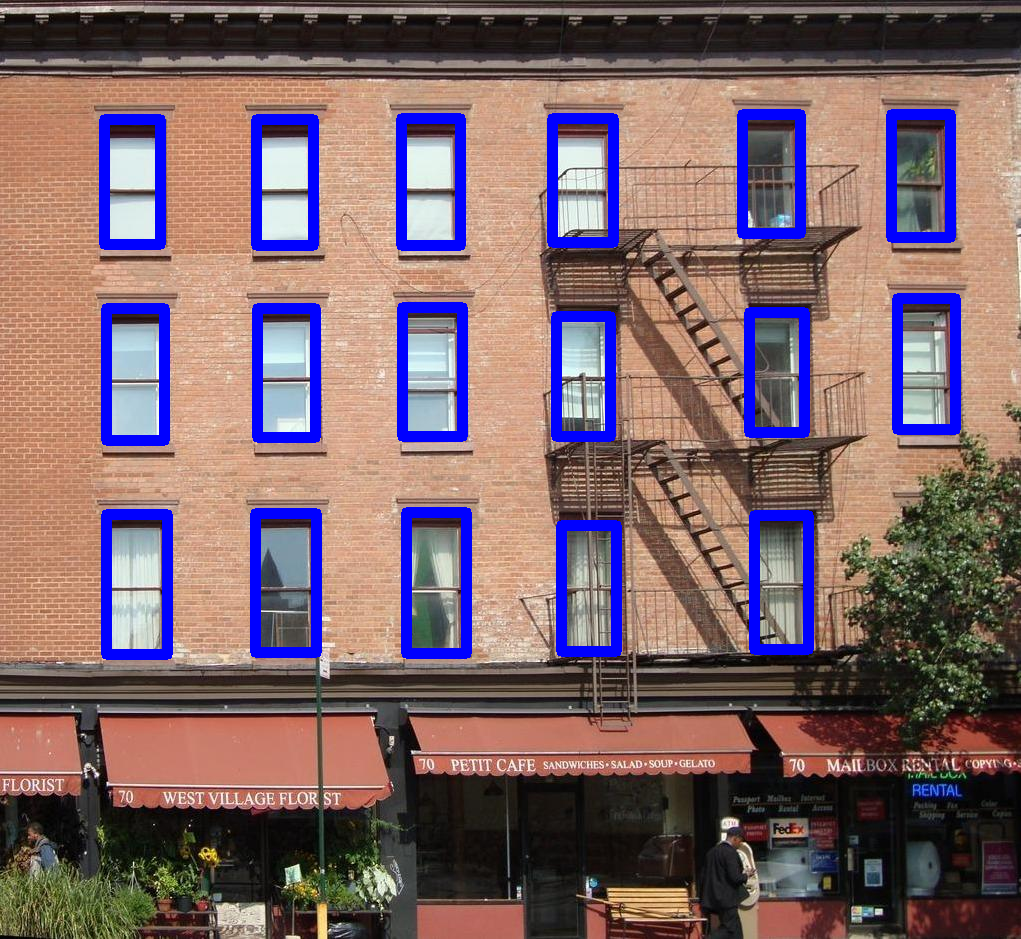} &
    \includegraphics[height=\fixedheight, width=0.32\linewidth, keepaspectratio]{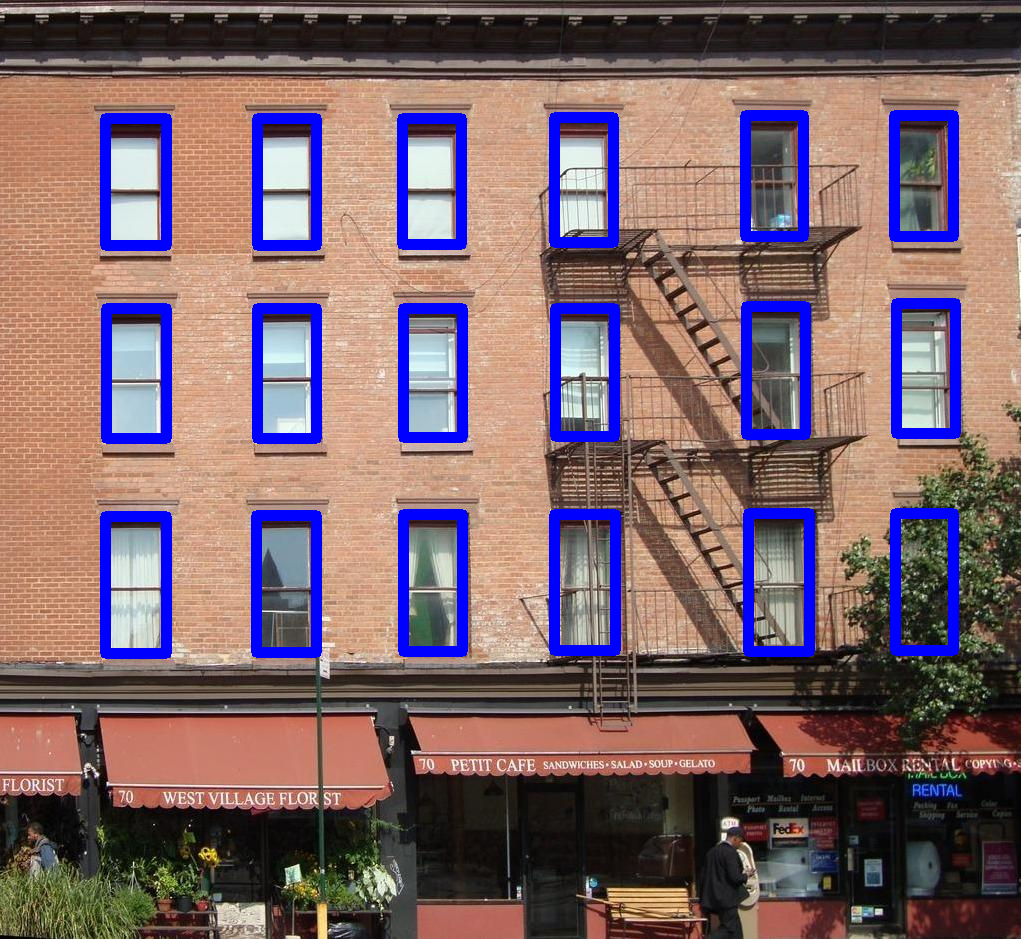} \\
    
    \includegraphics[height=\fixedheight, width=0.32\linewidth, keepaspectratio]{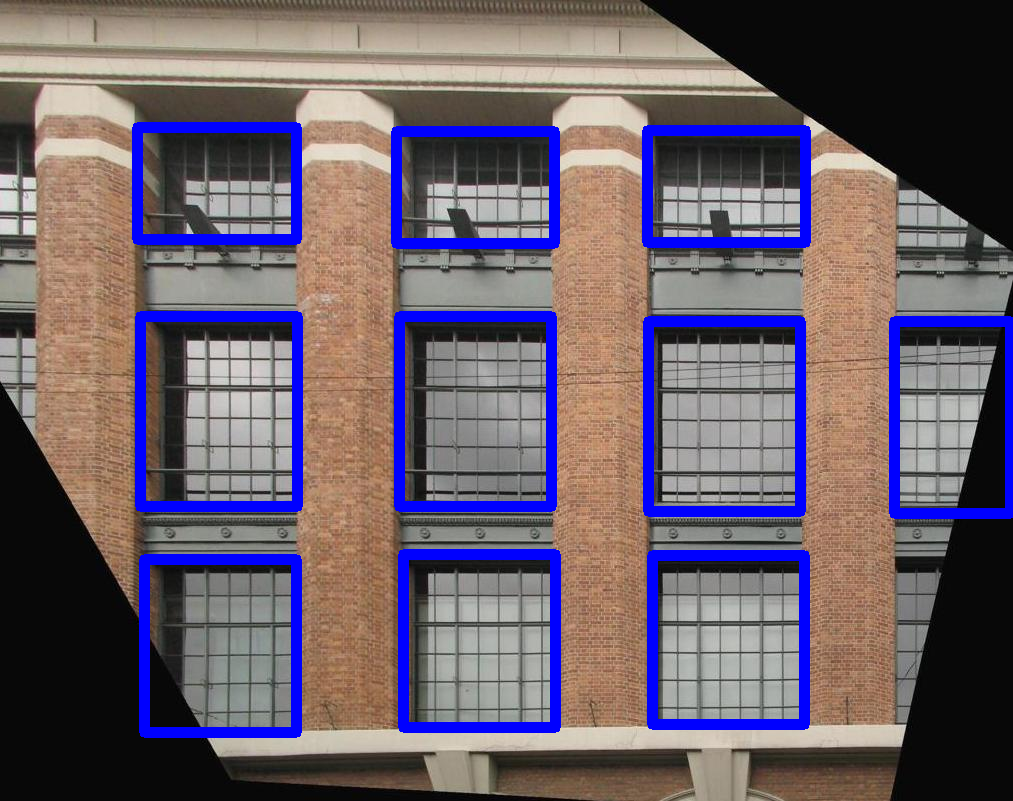} &
    \includegraphics[height=\fixedheight, width=0.32\linewidth, keepaspectratio]{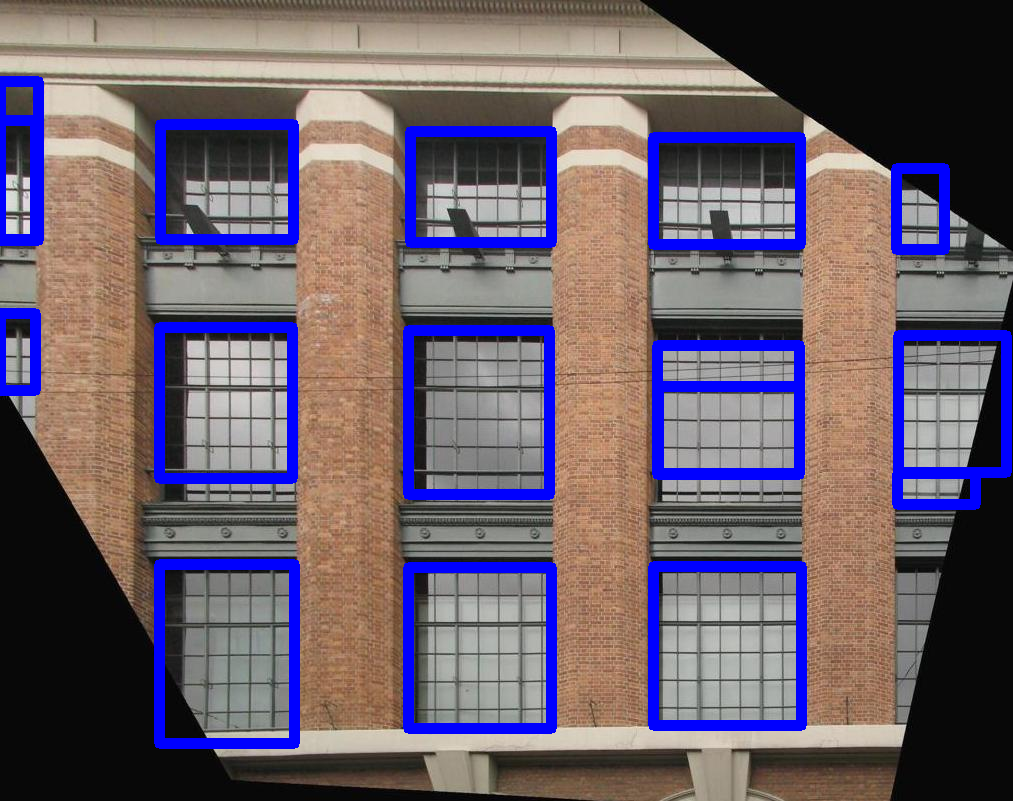} &
    \includegraphics[height=\fixedheight, width=0.32\linewidth, keepaspectratio]{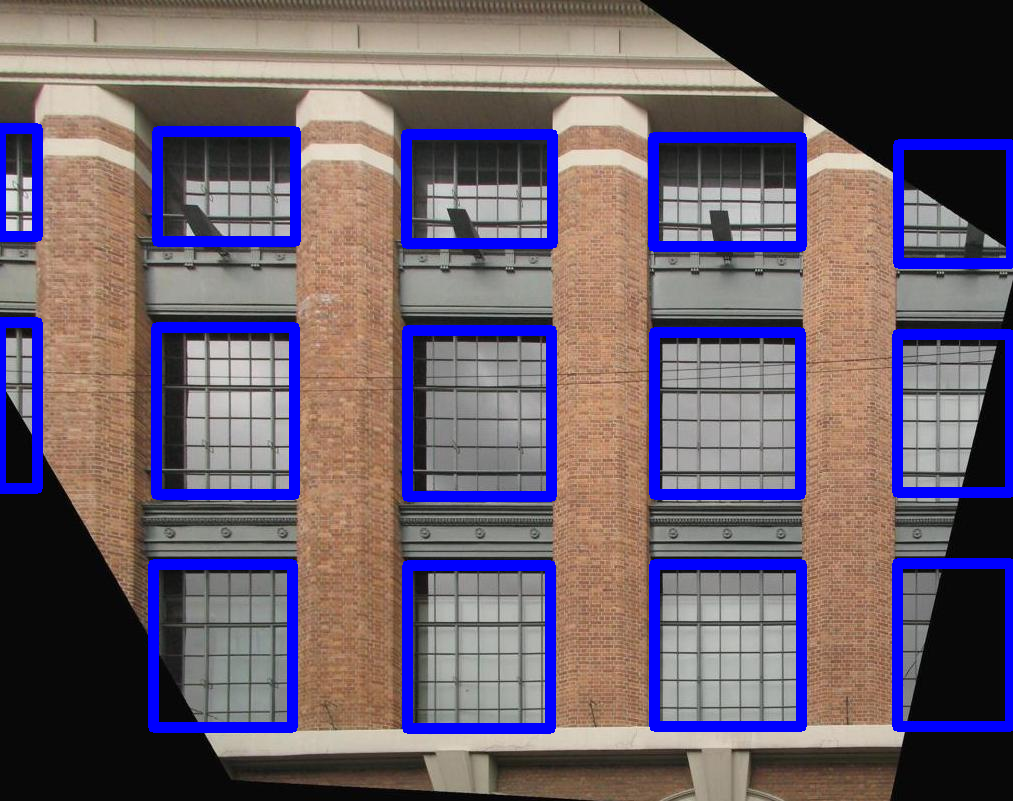} \\
    
    \includegraphics[height=\fixedheight, width=0.32\linewidth, keepaspectratio]{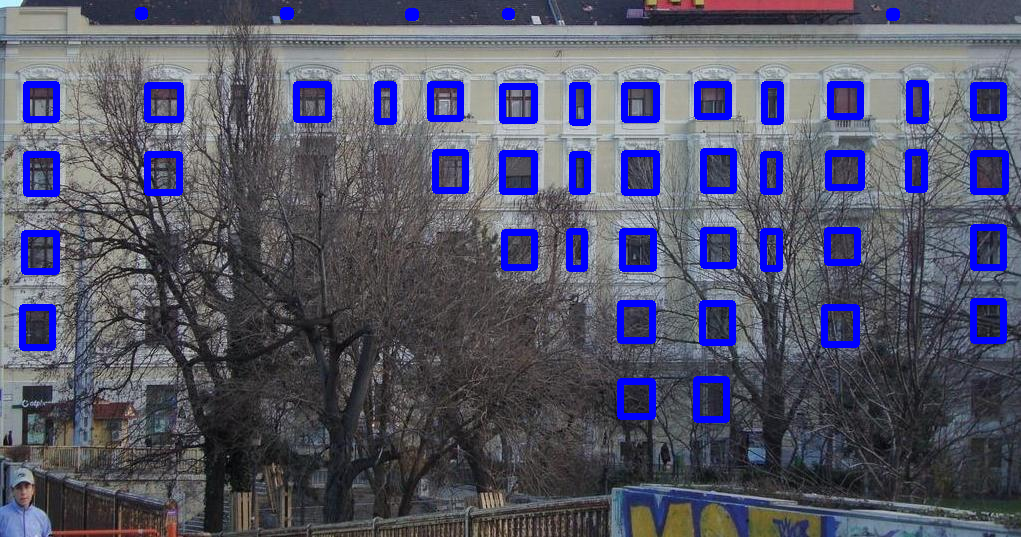} &
    \includegraphics[height=\fixedheight, width=0.32\linewidth, keepaspectratio]{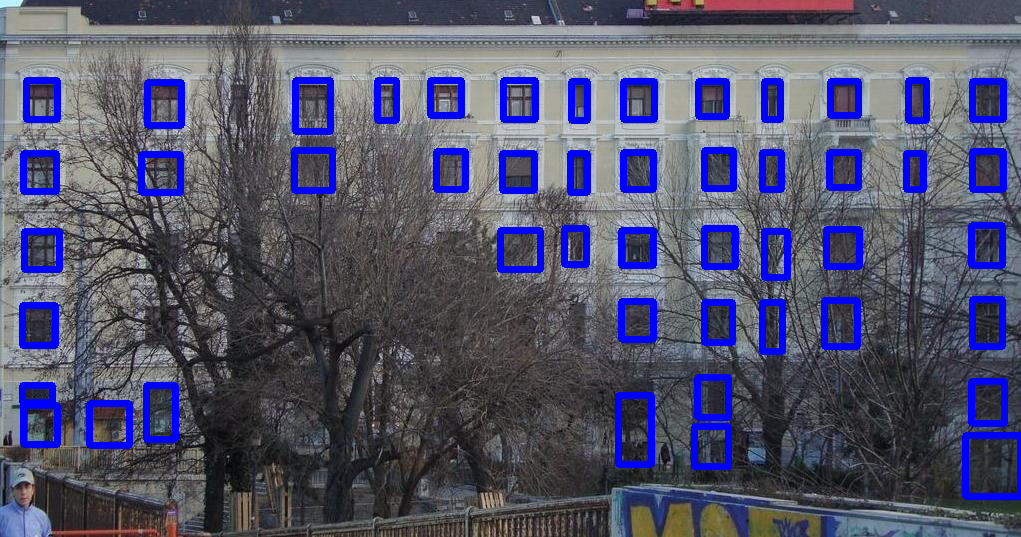} &
    \includegraphics[height=\fixedheight, width=0.32\linewidth, keepaspectratio]{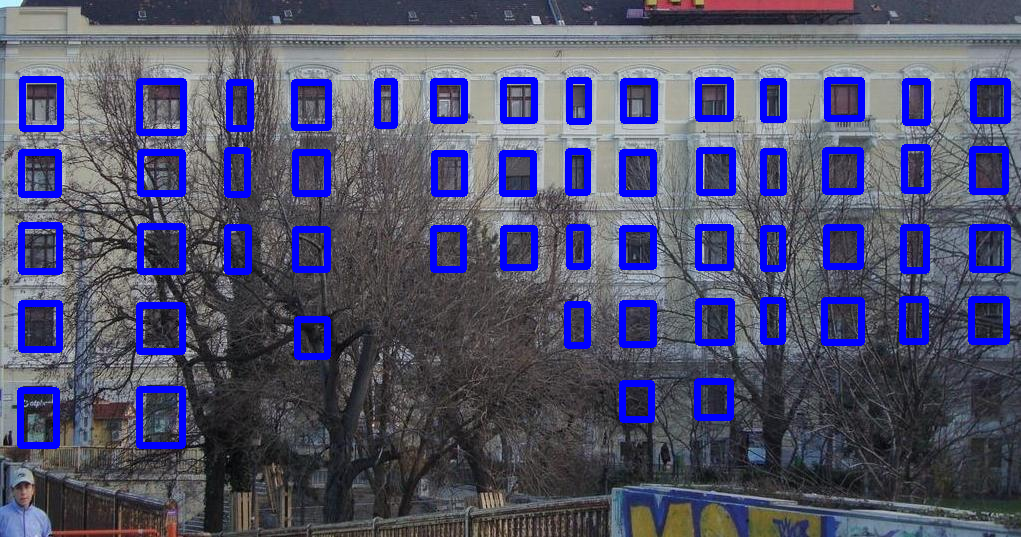} \\
  \end{tabular}
\end{minipage}%
\hfill
\begin{minipage}[t]{0.32\textwidth}
  \centering
  \begin{tabular}{ccc}
    \small\textbf{GT} & \small\textbf{Baseline} & \small\textbf{Ours} \\
    \includegraphics[height=\fixedheight, width=0.32\linewidth, keepaspectratio]{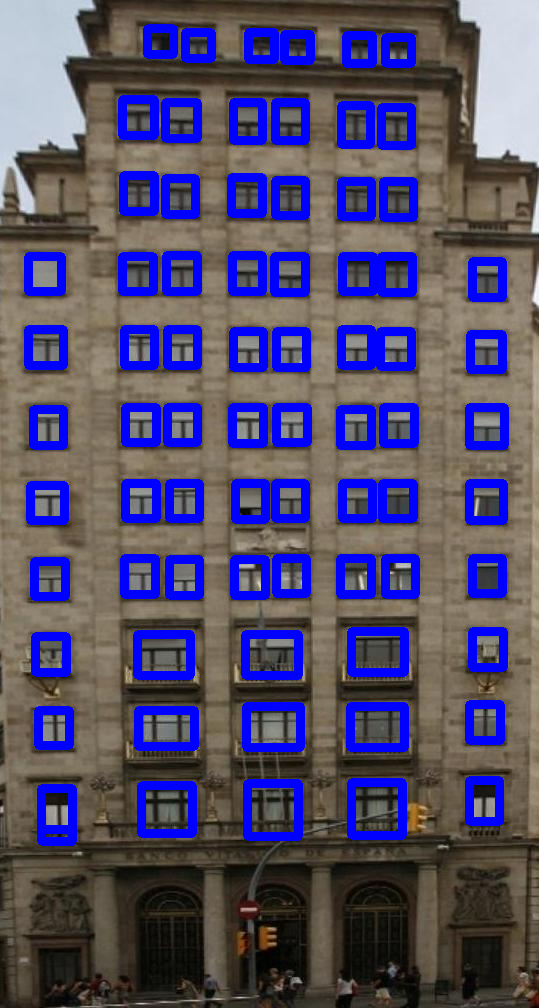} &
    \includegraphics[height=\fixedheight, width=0.32\linewidth, keepaspectratio]{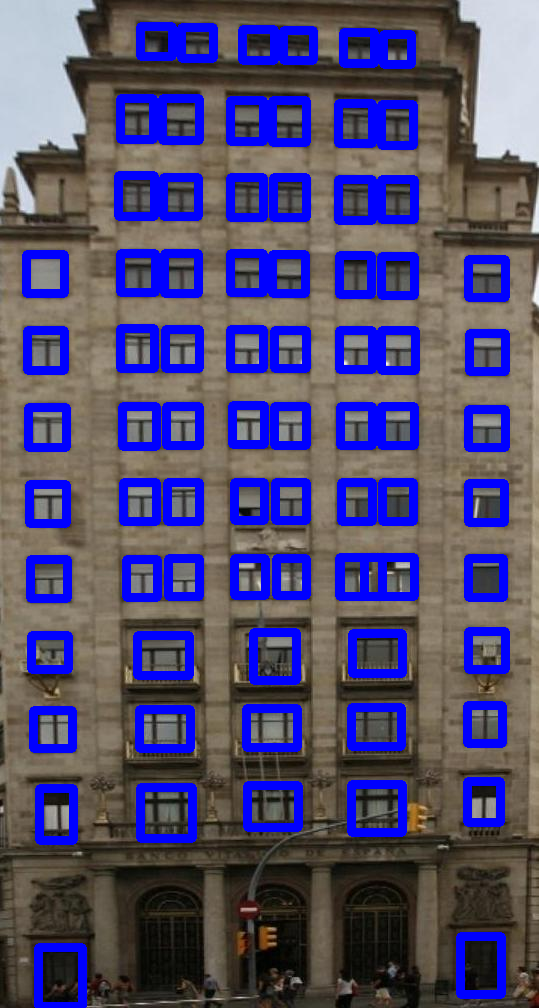} &
    \includegraphics[height=\fixedheight, width=0.32\linewidth, keepaspectratio]{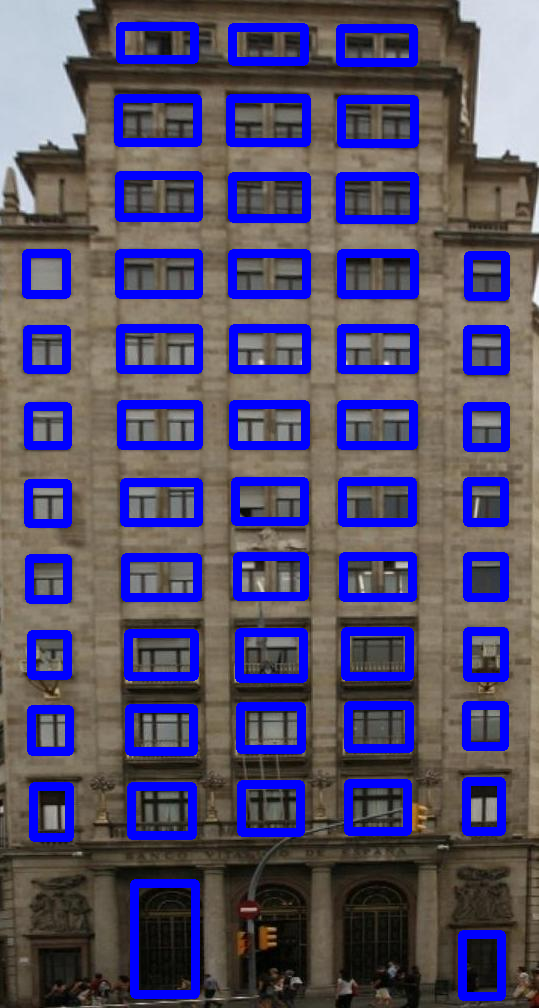} \\

    \includegraphics[height=\fixedheight, width=0.32\linewidth, keepaspectratio]{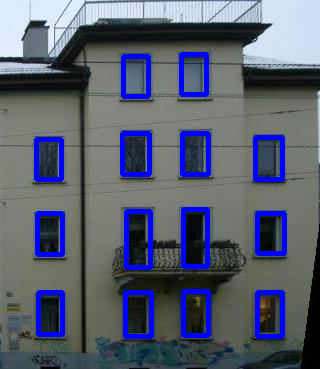} &
    \includegraphics[height=\fixedheight, width=0.32\linewidth, keepaspectratio]{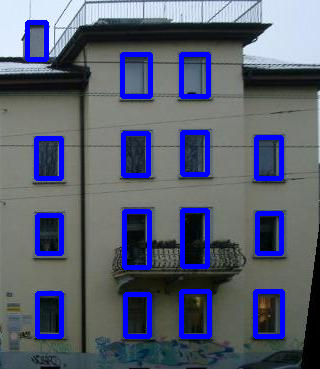} &
    \includegraphics[height=\fixedheight, width=0.32\linewidth, keepaspectratio]{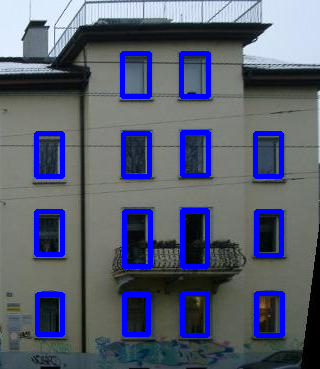} \\

    \includegraphics[height=\fixedheight, width=0.32\linewidth, keepaspectratio]{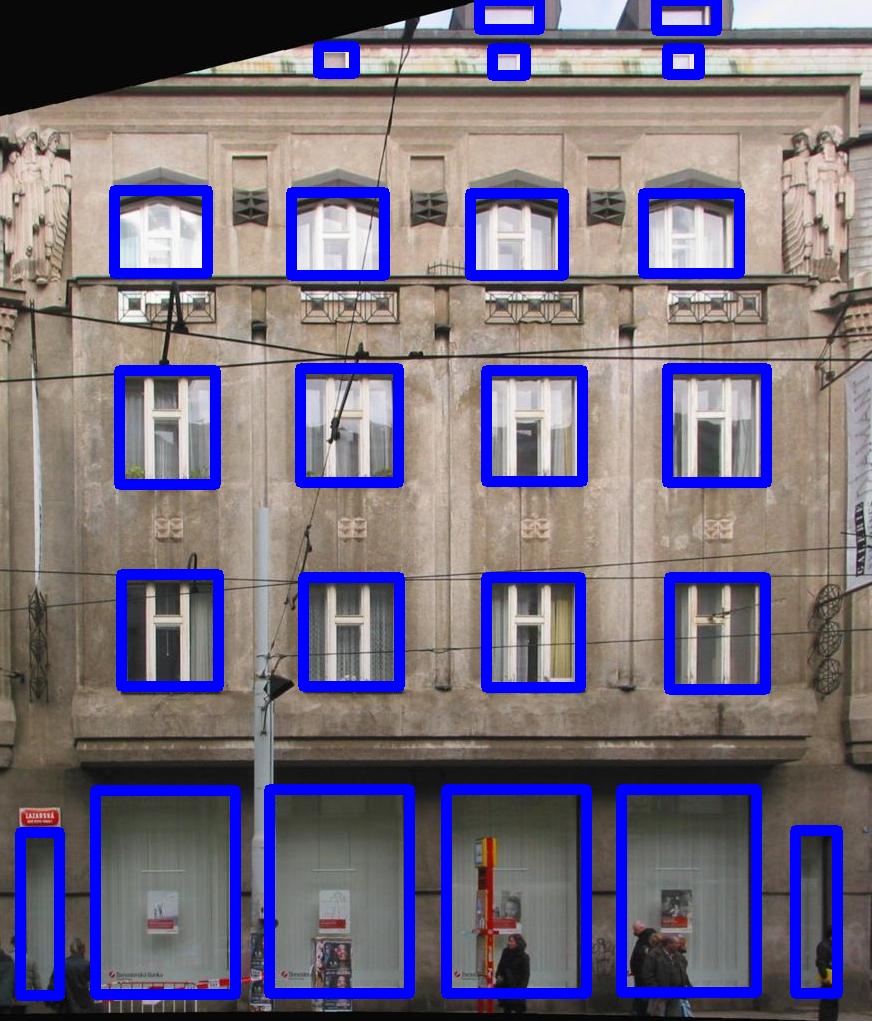} &
    \includegraphics[height=\fixedheight, width=0.32\linewidth, keepaspectratio]{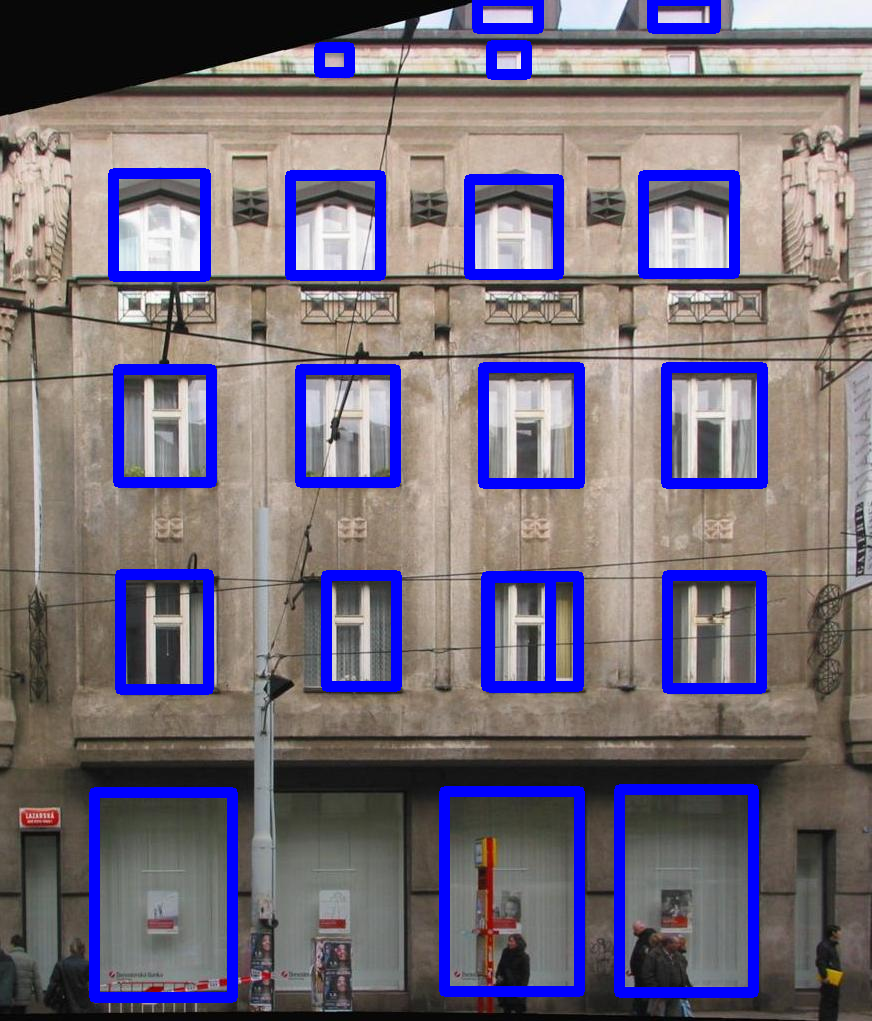} &
    \includegraphics[height=\fixedheight, width=0.32\linewidth, keepaspectratio]{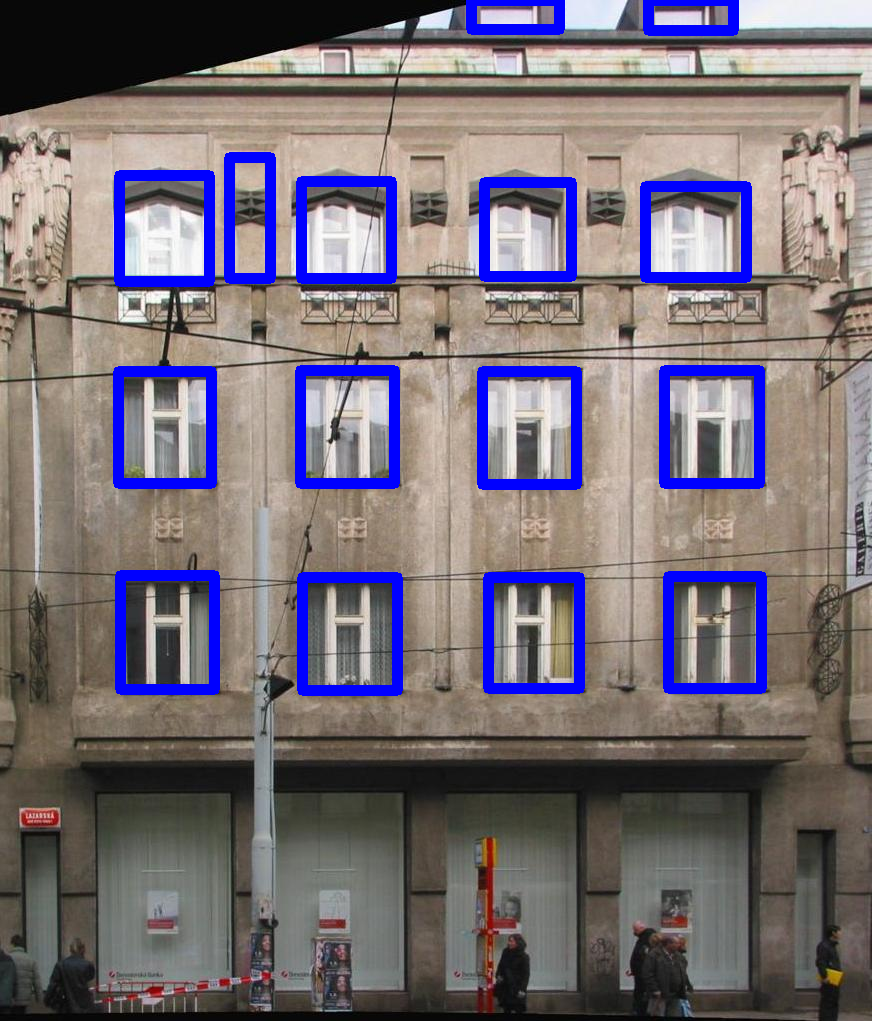} \\

    \includegraphics[height=\fixedheight, width=0.32\linewidth, keepaspectratio]{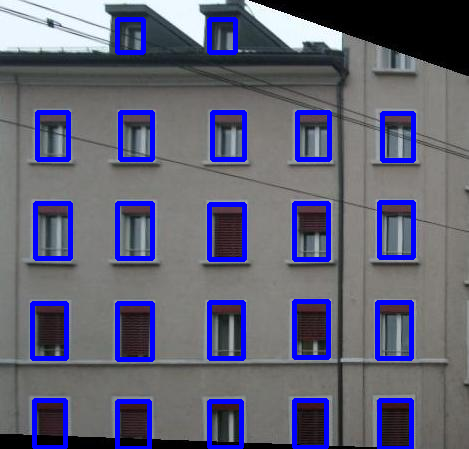} &
    \includegraphics[height=\fixedheight, width=0.32\linewidth, keepaspectratio]{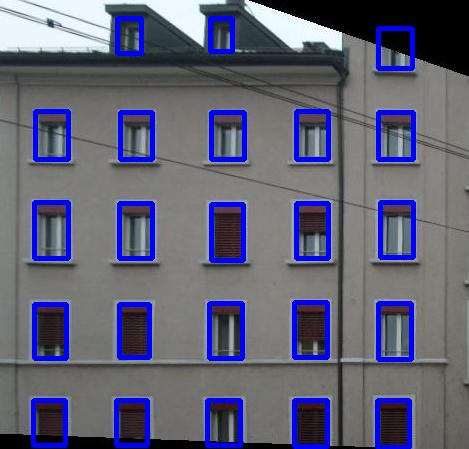} &
    \includegraphics[height=\fixedheight, width=0.32\linewidth, keepaspectratio]{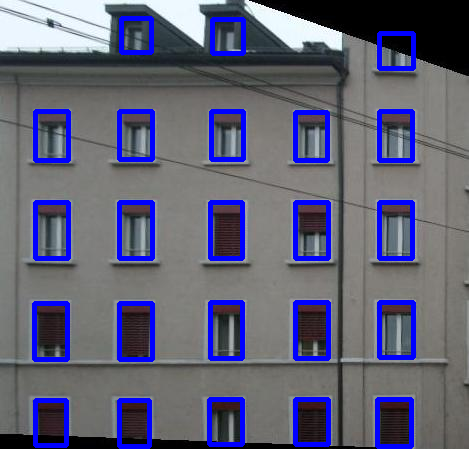} \\
  \end{tabular}
\end{minipage}%
\hfill
\begin{minipage}[t]{0.32\textwidth}
  \centering
  \begin{tabular}{ccc}
    \small\textbf{GT} & \small\textbf{Baseline} & \small\textbf{Ours} \\
    \includegraphics[height=\fixedheight, width=0.32\linewidth, keepaspectratio]{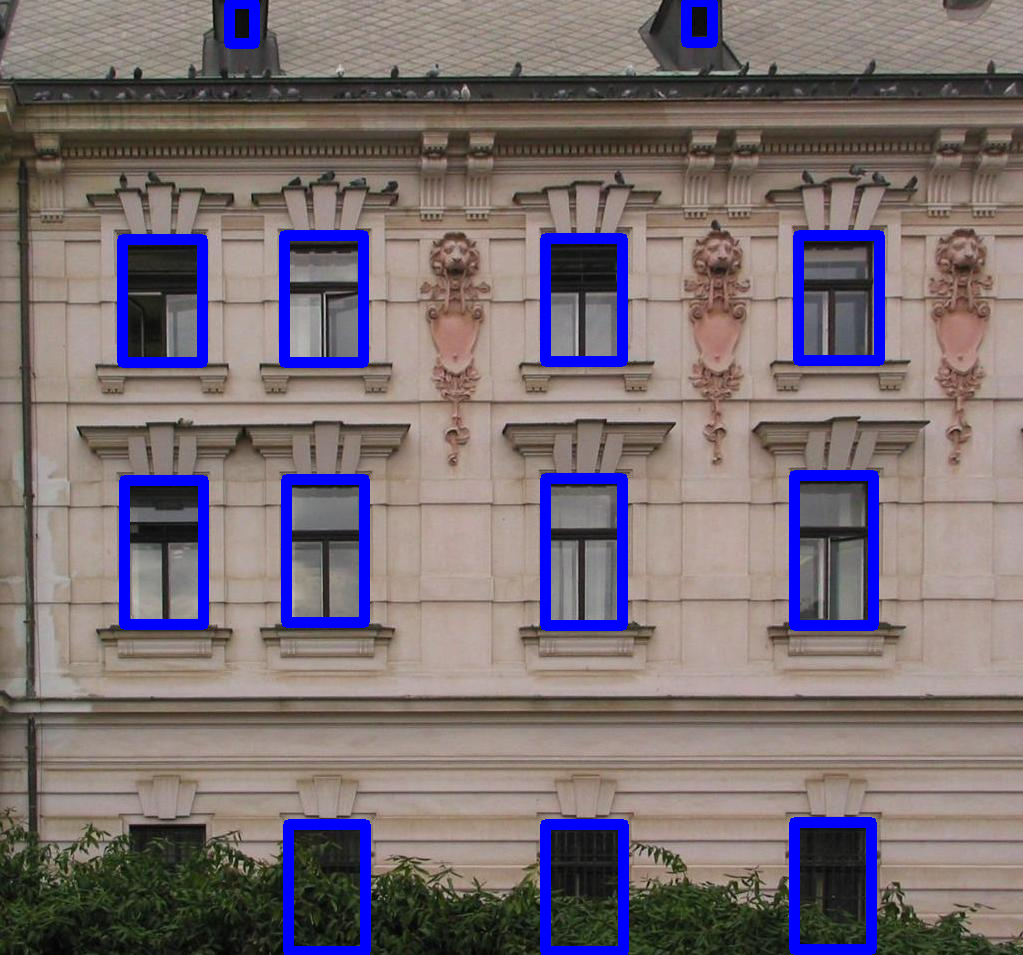} &
    \includegraphics[height=\fixedheight, width=0.32\linewidth, keepaspectratio]{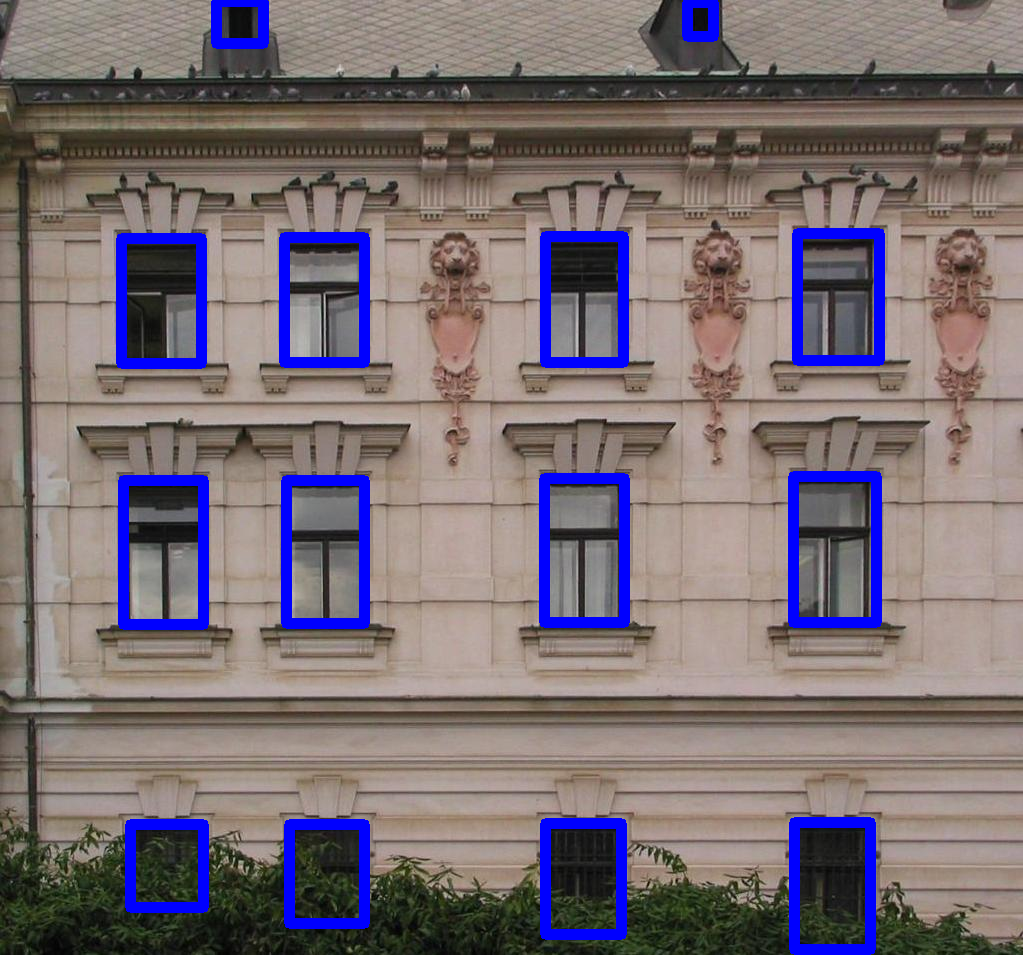} &
    \includegraphics[height=\fixedheight, width=0.32\linewidth, keepaspectratio]{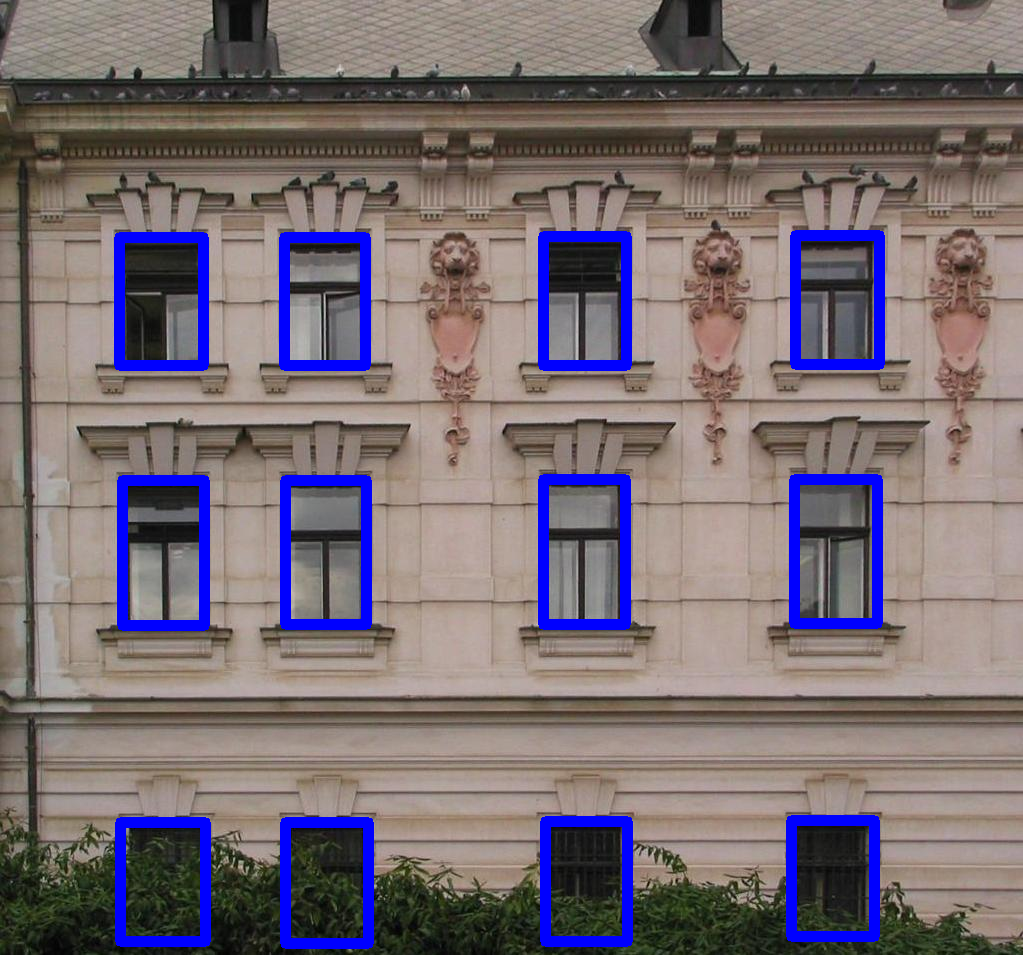} \\

    \includegraphics[height=\fixedheight, width=0.32\linewidth, keepaspectratio]{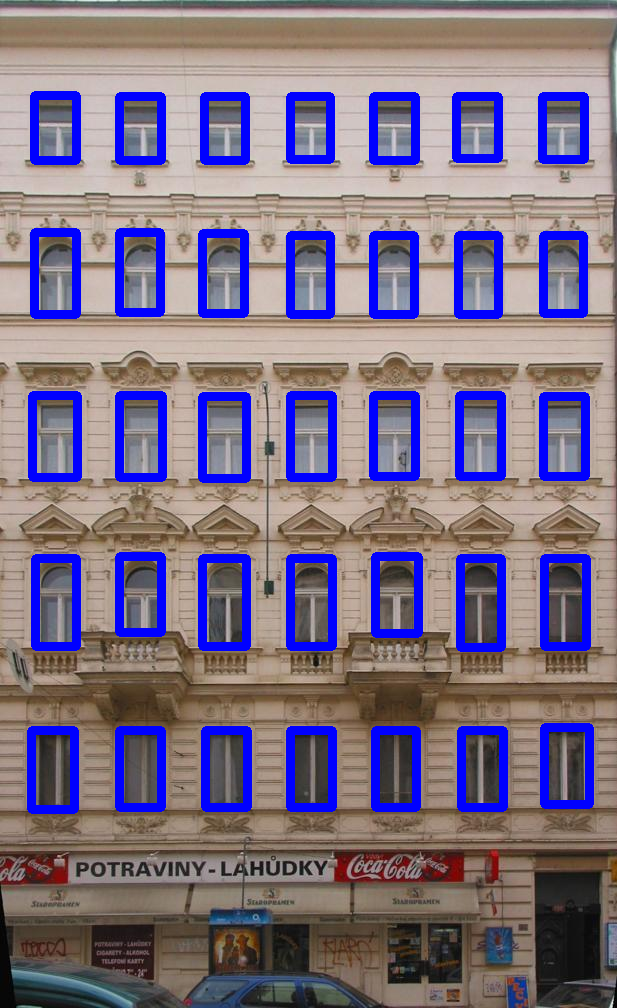} &
    \includegraphics[height=\fixedheight, width=0.32\linewidth, keepaspectratio]{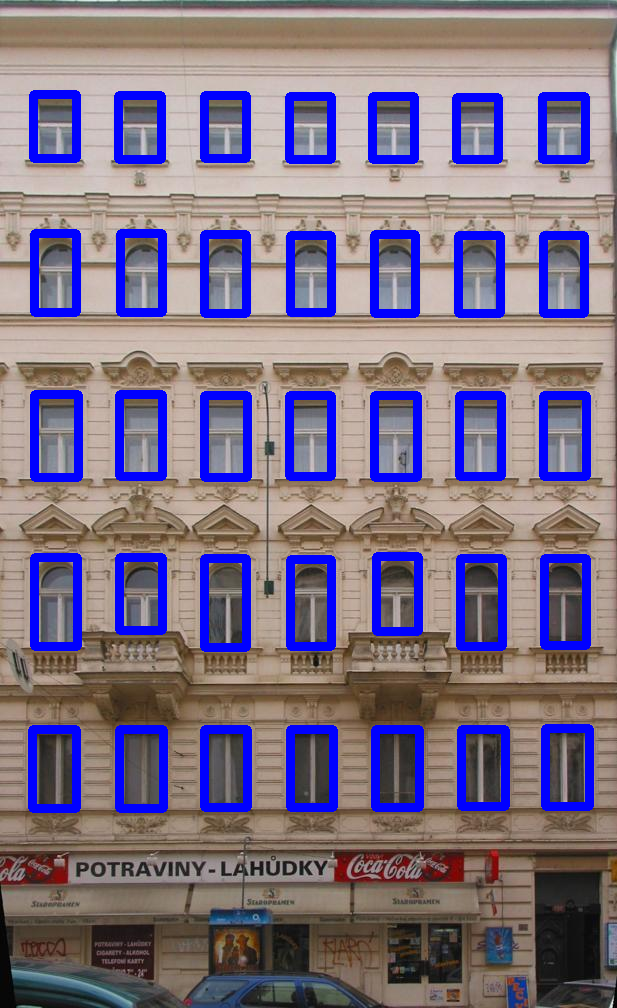} &
    \includegraphics[height=\fixedheight, width=0.32\linewidth, keepaspectratio]{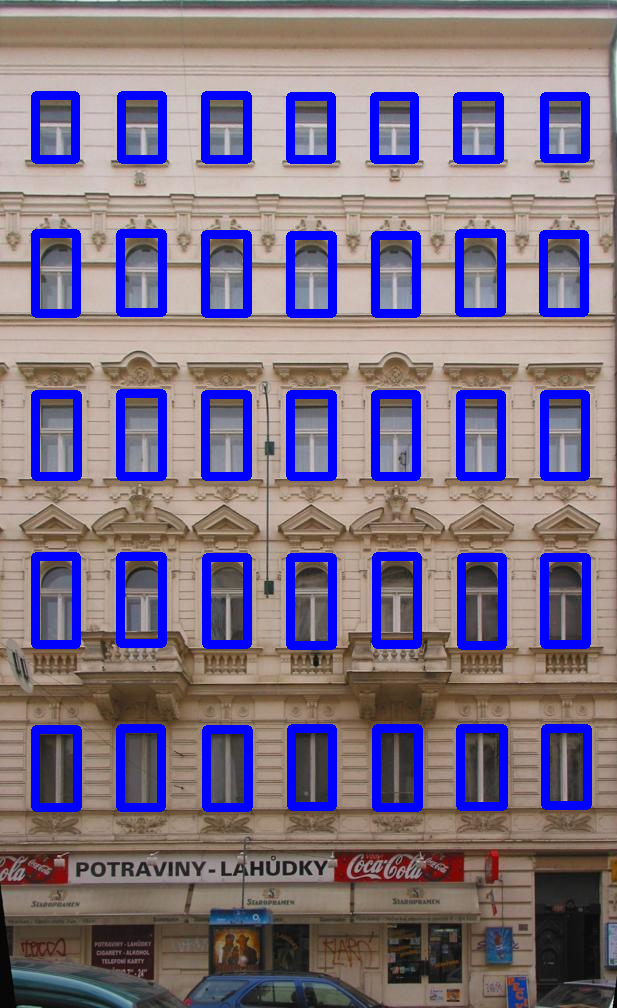} \\

    \includegraphics[height=\fixedheight, width=0.32\linewidth, keepaspectratio]{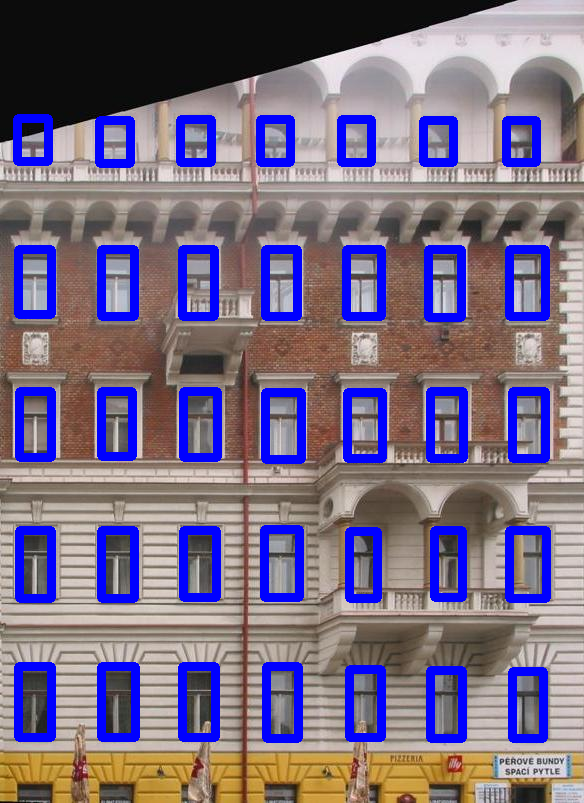} &
    \includegraphics[height=\fixedheight, width=0.32\linewidth, keepaspectratio]{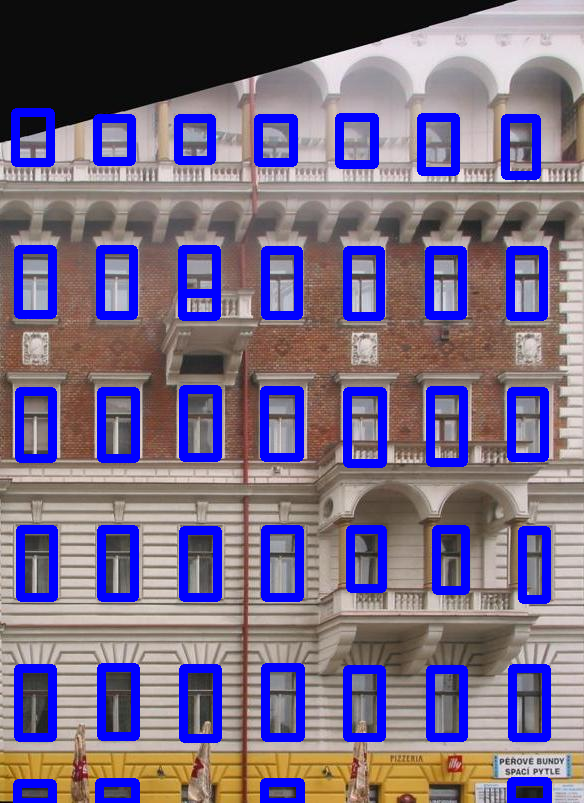} &
    \includegraphics[height=\fixedheight, width=0.32\linewidth, keepaspectratio]{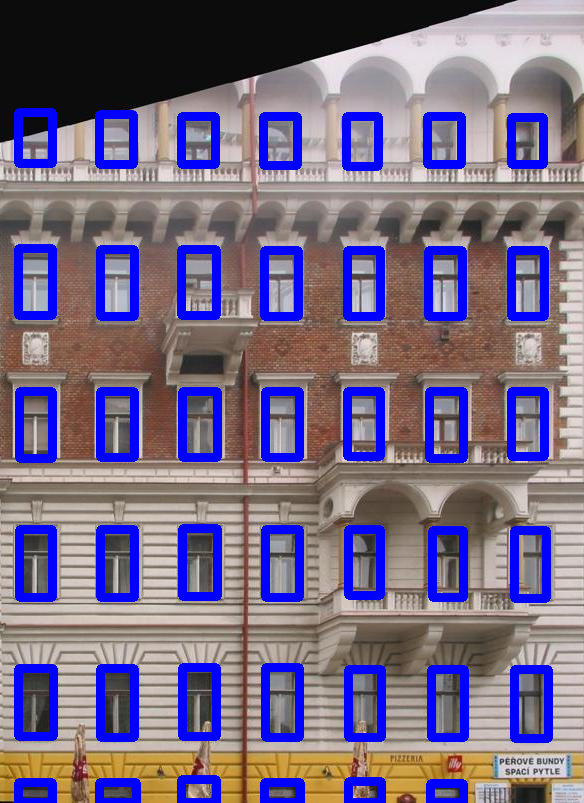} \\

    \includegraphics[height=\fixedheight, width=0.32\linewidth, keepaspectratio]{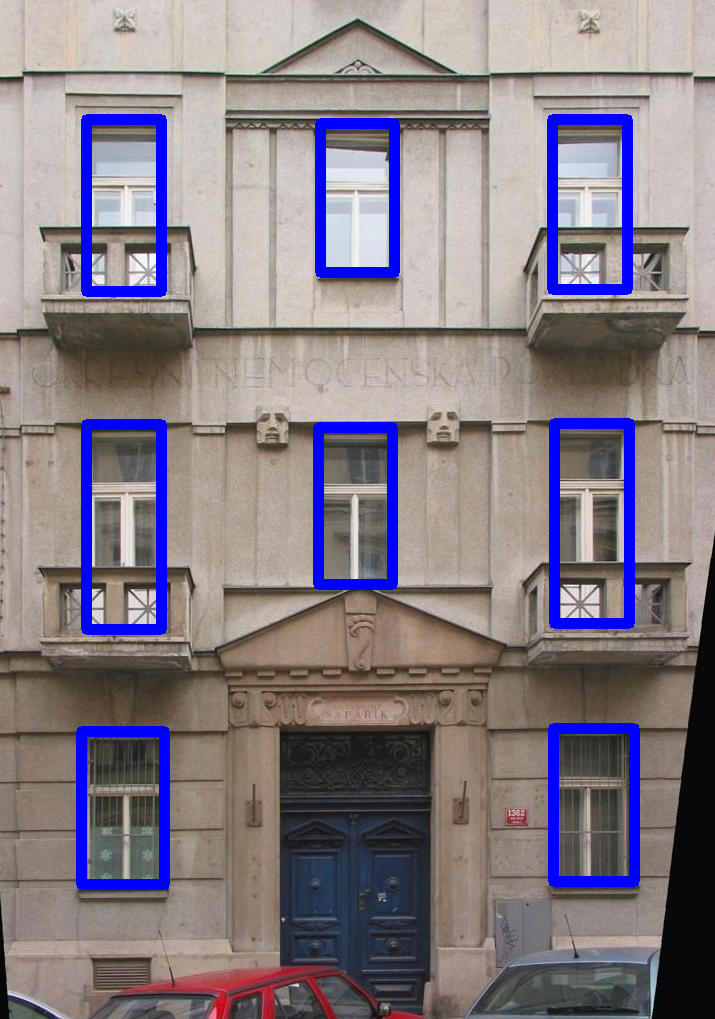} &
    \includegraphics[height=\fixedheight, width=0.32\linewidth, keepaspectratio]{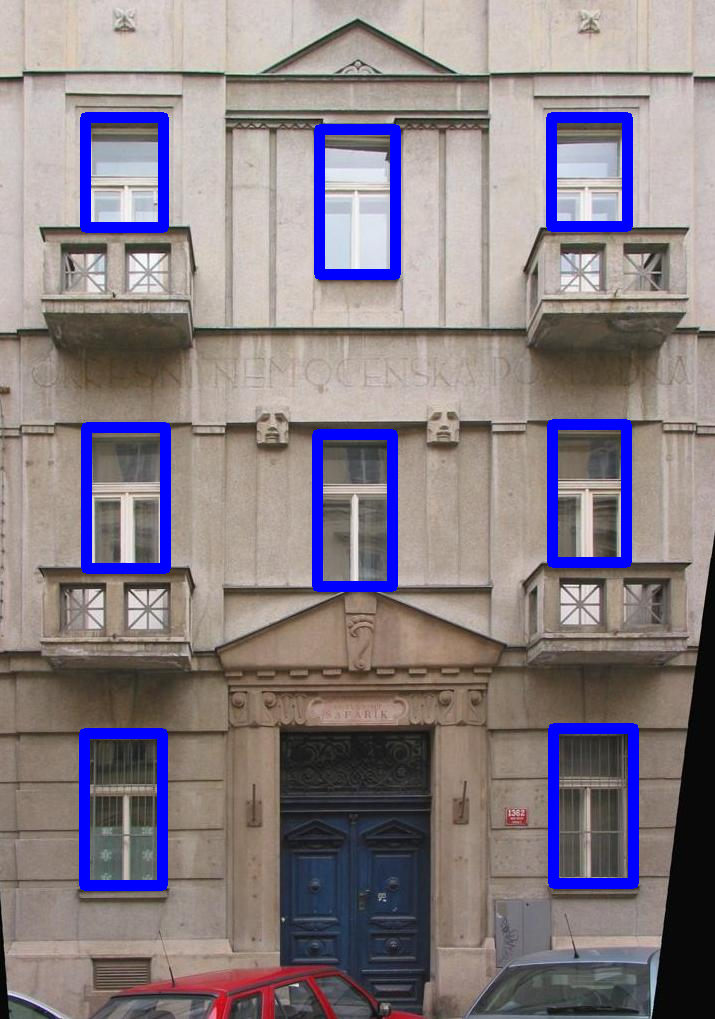} &
    \includegraphics[height=\fixedheight, width=0.32\linewidth, keepaspectratio]{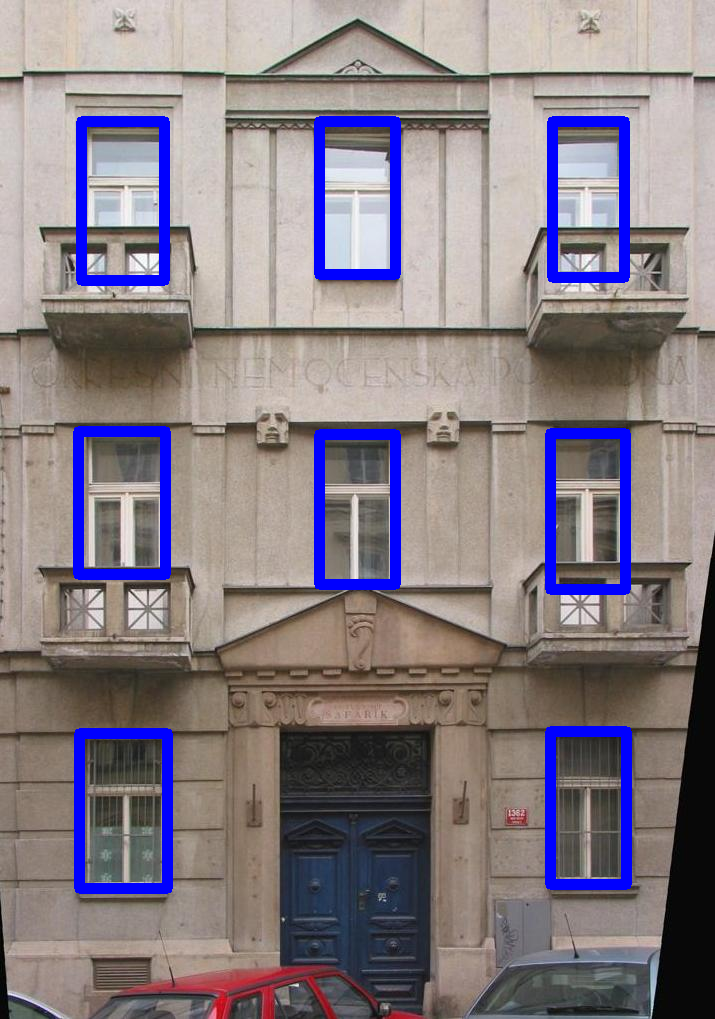} \\
  \end{tabular}
\end{minipage}

\caption{Qualitative results mosaic. Each row shows the Ground Truth (left), Baseline prediction (middle), and our Aligned prediction (right). The alignment loss (Ours) effectively restores grid-like regularity.}
\label{img:mosaic}
\end{figure*}

\noindent\textbf{Quantitative results. }
We evaluate multiple models based on their values of threshold $T$ and weight $W$. \del{The full quantitative results are listed in Table~\ref{tab:results} and visualized in Figure~\ref{fig:SVD_W} and \ref{fig:SVD_mAP}. }Results \rev{(see Figure~\ref{fig:SVD_W} and \ref{fig:SVD_mAP})} show that augmenting YOLO with our loss term lowers the value of the SVD-based metric, indicating higher regularity. By changing the $W$ value, we can control the prevalence of this effect. On the other hand, increasing \del{the weight }$W$ lowers the mAP score. For moderate values of $W$, one can decrease the SVD-based metric significantly without a substantial drop in mAP score, achieving a good balance between staying true to the original image while enforcing a structured result. However, further decreasing the SVD-based metric by manipulating $W$ comes with a steep drop in mAP score.

To further evaluate the loss term's performance, \del{three}\rev{five} values of threshold $T$ were tested. In the case of $T=9$\rev{, $T=10$ }and $T=12$, the trade-off between the SVD-based metric and the mAP score is \del{much }better than for $T=6$ and $T=7$. Also, for \del{$T=6$}\rev{small values of the threshold} and a high value of weight $W=1.0$, the SVD-based metric increases to \del{well }over $100\%$ (relative to the baseline), likely due to over-constraining. \del{On the other hand, there is no significant difference between $T=9$ and $T=12$ in terms of the trade-off between metrics.} \rev{Therefore, too small values of the threshold $T$, which controls the range of the alignment term, lead to worse model performance.}

\noindent\textbf{Qualitative results. } 
Figure~\ref{img:mosaic} shows example detections from our model compared to the dataset ground truth and baseline results. The model used was trained with $T=9$ and $W=0.5$. For better readability, we highlight only the window bounding boxes.

In the first column, one can see that the model successfully detected more windows that were partially or fully obstructed by trees, and even detected windows in missing parts of images (black triangles in the third row). The second column of Figure~\ref{img:mosaic} shows improvement over the baseline model and the dataset annotations in terms of window alignment. In all four examples, windows that were previously misaligned due to perspective were aligned with the rest of the windows by our model. This corresponds much better with the true facade structure. The top right example shows that windows visible in the bottom part of the facade are incorrectly detected by the baseline model in terms of their sizes, while the CMP ground truth misses one of the windows. Our model predicts all of them and estimates their sizes to be consistent. The other three images in the last column show that our model is better at approximating the sizes of windows partially obstructed by balconies. In the third image of the last column, the top row of windows also gets aligned with other rows, eliminating the influence of perspective.

\rev{If the $W$ value is sufficiently large, it can sometimes lead to missed detections. In the third example of the second column, both very large and very small windows were not detected. A similar effect can be observed in the top right example, where small windows near the top of the image were also missed by the model.}

\vspace{-1mm}
\section{Limitations and Conclusions}
\vspace{-2mm}

\noindent\textbf{Limitations.}
While our method effectively improves structural consistency, it introduces specific constraints. First, there is an inherent trade-off between geometric regularity and detection accuracy. Increasing the alignment weight $W$ improves regularity \del{(lowering the SVD score) }but can degrade mAP@0.5, particularly at higher weights where the model prioritizes alignment over pixel-level evidence. \del{Second, }Especially, by penalizing edge discrepancies, the loss implicitly favors equal widths or heights within aligned pairs. In facades with mixed window types within the same row or column, this can lead to unintended stretching or shrinking of bounding boxes. \rev{Second, the effectiveness of the alignment loss term is limited.} The method relies on the detector finding at least two elements to form a pair; consequently, singleton classes or sparse elements that rarely align receive little benefit. \rev{Likewise, a reasonable choice of $W$ and $T$ values can align bounding boxes only if the perspective distortion is relatively small. This is intentional: high perspective influence is indistinguishable from a structural misalignment of facade elements. Also}, \del{Finally, }in cases of severe occlusion where the detector fails to propose any candidates in a region, the alignment loss cannot recover the missing structure, as it operates only on positive predictions. \rev{Finally, the $T$ parameter is defined as an absolute number of pixels, which makes the alignment term's performance dependent on the image resolution.}

\noindent\textbf{Conclusions.}
We presented a detector-side regularization strategy that encourages grid-consistent facade detections by augmenting YOLOv8 training with a pairwise alignment loss. Experiments on the CMP dataset demonstrate that this approach improves structural regularity, as quantified by an SVD-based metric, with a controllable trade-off against standard detection accuracy. Overall, our method provides a lightweight mechanism for producing more grid-consistent detections suitable for procedural reconstruction, while preserving the efficiency of the standard inference pipeline.

\noindent\textbf{Future Work.}
Future avenues include extending alignment regularization to 3D layouts for inverse procedural modeling and integrating these grid-consistent detections as reliable anchors for explicit structure completion in occluded regions~\cite{fan2014structure}.

\vspace{-1mm}
\section{Acknowledgments}
\vspace{-2mm}

\rev{This research was carried out with the support of the High Performance Computing Center at Faculty of Mathematics and Information Science Warsaw University of Technology. Generative AI tools were used for brainstorming and refining the manuscript. All scientific content was produced entirely by the authors.}

\bibliographystyle{eg-alpha-doi} 
\bibliography{ShortPaper}

\end{document}